\documentclass[fleqn,usenatbib]{rasti}

\usepackage{newtxtext,newtxmath}

\usepackage[T1]{fontenc}

\DeclareRobustCommand{\VAN}[3]{#2}
\let\VANthebibliography\thebibliography
\def\thebibliography{\DeclareRobustCommand{\VAN}[3]{##3}\VANthebibliography}

\usepackage{graphicx}	
\usepackage{amsmath}	

\usepackage{subcaption}
\usepackage{booktabs}

\title[Operational Range Bounding in Spectroscopy]{Operational Range Bounding in Spectroscopy: A Safety Cage Framework for Machine Learning Models}

\author[N. Grens et al.]{
Nikki Grens,$^{1, 2}$
Lu{\'i}s F. Sim{\~o}es,$^{1}$\thanks{E-mail: luis.simoes@mlanalytics.ai}
Kai Hou Yip$^{1,3}$
and Theresa Lueftinger$^{4}$
\\
$^{1}$ML Analytics, 2660-329 Lisbon, Portugal\\
$^{2}$Department of Space Engineering, Delft University of Technology, 2600 AA Delft, The Netherlands\\
$^{3}$Department of Physics, King’s College London, University of London, WC2R 2LS, London, United Kingdom\\
$^{4}$European Space Research and Technology Centre (ESTEC), European Space Agency (ESA), 2201 AZ Noordwijk, The Netherlands
}

\date{Accepted XXX. Received YYY; in original form ZZZ}

\pubyear{\the\year{}}
 
\begin{document}
\label{firstpage}
\pagerange{\pageref{firstpage}--\pageref{lastpage}}
\maketitle

\begin{abstract}
Ensuring the reliability of black-box machine learning models in safety-critical space missions remains a significant challenge, particularly when ground-truth is unavailable for validation. Although machine learning models offer a powerful means to augment standard pipelines by extracting transmission spectra from complex exoplanetary light curves, their susceptibility to unmodelled instrument anomalies, stellar activity, and domain shifts introduces unquantified risks. This study evaluates a modular safety cage architecture that operates as a parallel monitoring layer to assess the validity of a prediction without modifying the underlying estimator. By monitoring different runtime indicators, including uncertainty quantification, out-of-domain detection, and influence functions, the framework constrains the model’s operational domain to a verified region. A controlled evaluation is conducted under both in-domain and cross-domain conditions, using datasets from the 2019 and 2021 editions of the Ariel Data Challenges. The results reveal that model failure is multifaceted and that no single indicator captures all failure modes, demonstrating the need for indicator fusion. The application of safety-driven rejection strategies shows that a modest 20\% reduction in data coverage results in error reductions between 45\% and 65\% across different domains and evaluation metrics. Using a formalised coverage-risk framework, a systematic analysis of indicator combinations is performed to identify configurations that maximise risk-ranking accuracy and optimise the trade-off between data coverage and scientific performance. Safety cages provide a transparent mechanism for detecting unreliable predictions and represent a critical step towards the safe deployment of data-driven models in scientific applications, such as astrophysics, where ground truth is seldom available.
\end{abstract}



\section{Introduction}
Characterisation of exoplanetary atmospheres represents a fundamental shift in understanding planetary formation and evolution, with the European Space Agency’s (ESA) Ariel mission designed to conduct the first large-scale chemical census of approximately 1000 exoplanets \citep{tinetti2021arielenablingplanetaryscience}. Central to this mission is transit spectroscopy, a technique that captures a planet's chemical fingerprint by observing stellar light filtered through its atmosphere when the planet passes in front of its host star \citep{tinetti2018chemical}. During this transit event, the planet obstructs a portion of the stellar disc, causing a measurable drop in the total observed light. As stellar photons pass through the atmosphere, specific chemical elements and molecules absorb light at characteristic wavelengths. By decomposing the light into a transmission spectrum, the depth of the transit across different wavelengths can be measured, and the variations can be used to infer the atmospheric composition, temperature, and pressure of a world tens of parsecs away. In this context, the input signals are the observed time-series light curves, while the target outputs are the transmission spectra. Extracting these signals of interest remains extraordinarily challenging, as atmospheric signatures as faint as 10–100 parts per million (ppm) are frequently obscured by complex instrumental systematics (such as $1/f$ noise or persistence) and astrophysical noise (such as stellar activity).

Although machine learning (ML) models have been successful in extracting these subtle signals, their deployment in safety-critical space missions faces a significant challenge due to the lack of formal reliability guarantees \citep{AIsurvey, carleo2019machine}. In space-based applications where ground truth is rarely available and in situ measurements are not possible, traditional estimators lack automated mechanisms to verify whether unmodeled noise patterns have been misinterpreted as genuine atmospheric signatures \citep{doshi2017towards, jia2021physics}. To address this limitation, a safety cage architecture was designed to serve as a parallel monitoring layer \citep{ECSShandbook, DEEL}. Operating alongside the predictive model, this framework continuously validates input/output pairs at runtime without modifying the underlying estimator, constraining predictions to a verified region of competence \citep{safetycages}.

This work investigates whether runtime safety indicators, such as uncertainty estimates, ensemble disagreement, and activation anomalies, correlate with and can predict loss of model performance and degradation in transmission spectrum reconstruction when ground truth is unavailable. The Ariel Data Challenge (ADC) datasets from the 2019 and 2021 editions provide a controlled framework for investigating the safety cage architecture both in-domain and cross-domain \citep{yip_2025_15050868, nikolaou2023lessons}. By incorporating distinct noise profiles and instrument modelling assumptions, these datasets enable a rigorous analysis of model performance under data drift. Evaluating the framework across both operational domains makes it possible to determine if safety indicators can detect scientific degradation before significant reconstruction errors occur.

The contributions of this study are threefold. First, a modular safety cage architecture is introduced that integrates uncertainty quantification, out-of-domain detection, internal model monitoring, sensitivity quantification, and data attribution within a unified framework. Second, a controlled comparative evaluation of multiple safety indicators is conducted using the ADC19 and ADC21 datasets, benchmarking performance both in-domain conditions and under cross-domain scenarios involving data drift. Third, the operational benefits of rejection strategies are quantified through coverage-error analysis, demonstrating that safety-driven filtering can reduce catastrophic reconstruction errors while preserving scientifically valid predictions.

\section{Background Information}
ML models have achieved state-of-the-art performance in complex regression and inference tasks, yet their deployment in safety-critical systems remains challenging. Traditional performance metrics, such as mean squared error, provide aggregate measures of the model's accuracy on existing data for which ground-truth is available, but offer no guarantees regarding behaviour under novel or adverse conditions in the absence of ground truth. Certification standards in aerospace, automotive, and other high-assurance domains emphasise the need for clearly defined operational domains and runtime monitoring mechanisms capable of detecting deviations from validated behaviour \citep{ECSShandbook, DEEL, EASA, systematicliteraturereview, perez2024artificial, borg2018safelyenteringdeepreview}. In space missions, this requirement is particularly stringent \citep{tinetti2021arielenablingplanetaryscience}. When ground truth is unavailable, as in exoplanet atmospheric observations or other astronomical observations, incorrect predictions cannot be directly falsified at runtime. Consequently, reliability must be inferred from observable properties of the model and the data.

\subsection{Safety Cage Architecture and Operational Domains}
\begin{figure}
    \centering
    \includegraphics[width=\linewidth]{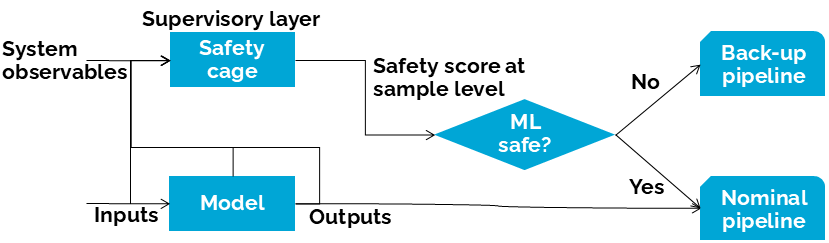}
    \caption{Block diagram of the safety cage architecture. The predictive model operates alongside a parallel supervisory layer that evaluates runtime safety indicators from the input, internal model states, outputs, and system observables. A gating mechanism then accepts or rejects each prediction based on whether it falls within the validated operational domain. Schematic adapted from \citet{ECSShandbook}. }
    \label{fig:SCarchitecture}
\end{figure}

The safety cage architecture has been proposed as a structured approach to integrate runtime monitoring into ML pipelines \citep{safetycages, ECSShandbook, DEEL, safeAutomotiveSoftware}. In this architecture, shown in \autoref{fig:SCarchitecture}, a parallel supervisory layer evaluates safety indicators derived from input data, internal model states, model outputs, or the model's operational context. Rather than altering the predictive model itself, the cage constrains its operational domain by accepting or rejecting predictions based on predefined criteria. Central to this framework is the notion of a validated operational domain. Inputs may fall into known–safe (where nominal operation proceeds), known–unsafe (which requires intervention or assessment using alternative models), unknown–safe (representing regions that should be characterised and potentially integrated into the validated domain), or unknown–unsafe (which must be rigorously detected and rejected) regions, depending on prior knowledge of the system and empirical validation \citep{DEEL}. The objective is to detect transitions into unsafe or uncharacterised regions before they lead to system-level failure. While the safety cage concept is increasingly discussed in certified ML systems \citep{ECSShandbook}, empirical validation of its effectiveness in scientific inference pipelines remains limited.

\subsection{Uncertainty and Distributional Shift}
\begin{figure}
    \centering
    \includegraphics[width=\linewidth]{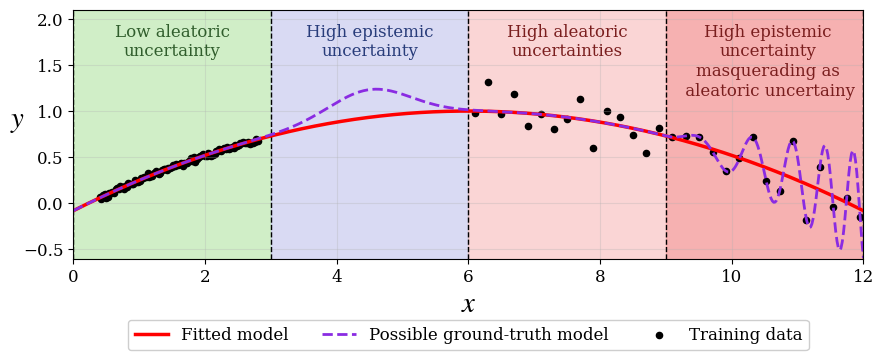}
    \caption{Schematic of aleatoric and epistemic uncertainty in machine learning models. The dashed purple line represents a theoretical ground truth model within the hypothesis space. While aleatoric uncertainty remains a constant property of data noise, epistemic uncertainty in segment four, characterised by high model variance, is reducible as the training data expands.}
    \label{fig:uncertainty_plot}
\end{figure}

Model reliability is impacted by different kinds of uncertainty, with literature distinguishing between aleatoric and epistemic. Aleatoric uncertainty arises from inherent stochastic variability in the data-generating process, whereas epistemic uncertainty reflects incomplete knowledge of the model, typically due to a limited amount of training data \citep{DEEL, systematicliteraturereview}. During the development phase of the predictive model, distinguishing between these sources is essential, as aleatoric variability may be irreducible, but epistemic uncertainty may signal that operations occur outside the training domain, allowing for the informed generation of new training data. \autoref{fig:uncertainty_plot} illustrates the contrast between these types of uncertainty, highlighting how regions with low data coverage can provide a distorted representation of the uncertainty at hand.

The preceding discussion assumed that the data-generating process remains stationary over time. However, in real-world applications, this is rarely the case. Systems evolve, instrumentation undergoes degradation, such as the Spitzer instrument entering its hot phase when its coolant was exhausted \citep{stauffer2007spitzer}, and environmental fluctuations, such as changing cosmic ray influxes on the James Webb Space Telescope due to the solar cycle \citep{jwst}, introduce novel phenomena. Consequently, the distribution of live data often shifts relative to the training data, a phenomenon called data drift \citep{amodei2016concreteproblemsaisafety, driftDetection}. For the Ariel mission, such shifts may manifest as evolving instrumental noise profiles, changing cosmic ray intensities, unforeseen sensor behaviour, or atmospheric features absent from training simulations. Additionally, it is also essential to distinguish between the specific modes of dataset shift, namely, covariate shift and concept drift. In covariate shift, the distribution of input features, such as signal-to-noise (SNR) ratios or detector temperatures, deviates from the training set. In concept drift, the fundamental mapping and relationship between the raw flux and the atmospheric parameters themselves change.

\subsection{Runtime Monitoring Mechanisms}
In practice, the safety cage architecture enforces runtime guards by continuously monitoring safety indicators to detect the aforementioned failure modes. When these indicators exceed predefined thresholds, the system triggers a fallback or safe state. Essentially, the cage acts as a protective isolation layer that enforces structural invariants over the ML model’s operation \citep{ECSShandbook}. Defining strict boundaries for acceptable behaviour ensures that any transition toward an unsafe state is intercepted, maintaining the safety of the predictable system even in the presence of uncertainty or model failure. This mechanism merely accepts or rejects predictions and does not improve the prediction itself. As no single indicator can capture every failure mode, a multi-layered approach is required to monitor the transition from high-fidelity simulations to real-world observations \citep{safetyperspectiveonautonomousdriving}. Researchers have proposed a variety of techniques to detect unsafe model behaviour, which can be grouped conceptually by the failure modes they address.

\subsubsection{Out-of-Domain Methods}
Out-of-Domain (OOD) detection aims to identify inputs that lie outside the support of the training distribution. Density estimation \citep{meinke2019towards, systematicliteraturereview}, distance-based metrics \citep{simpleunifiedframeworkOODandAdversarialattack}, space partitioning methods \citep{gu2019towards, systematicliteraturereview}, or anomaly detection models including isolation forests \citep{safetycages, chandola2009anomaly, liu2012isolation, liu2008isolation} are commonly employed to detect samples residing in low-probability regions of feature space. Such methods are effective for identifying unknown unknowns and inputs where the model lacks empirical grounding, signalling an increase in epistemic uncertainty.

\subsubsection{Uncertainty-Based Methods}
Current strategies for uncertainty quantification (UQ) encompass a diverse range of methodologies, including deep evidential regression, which provides a computationally efficient, single-pass estimation of both aleatoric and epistemic uncertainties \citep{deepevidentialregression}, and distribution-free frameworks such as conformal prediction. The latter transforms standard point predictions into interval predictions with rigorous finite-sample coverage guarantees, with the only assumption being data exchangeability \citep{shafer2008tutorial, angelopoulos2021gentle}.  Conformal prediction begins with a non-conformity measure, which quantifies the difference between the true and predicted output. By calculating these scores on a held-out calibration dataset, the system determines a quantile threshold. For any new test sample, a prediction interval is established by adding and subtracting this threshold from the point prediction, ensuring the true value falls within these bounds with a user-defined probability (e.g., 90\%). Complementary to these are methods that estimate model error directly through Quantile Regression (QR) \citep{QRKoenker, chung2021beyond}. Unlike global error estimates, QR generates sample-dependent intervals that provide a transparent measure of irreducible errors, such as photon noise or instrumental jitter, by learning conditional quantiles of the output distribution rather than a single point estimate. More broadly, these approaches represent model-based error estimation, where the predictive system generates uncertainty measures derived from its own residual structure or conditional distribution, providing a self-contained evaluation of the reliability of the output \citep{systematicliteraturereview}.

\subsubsection{Ensemble Methods}
To capture epistemic uncertainty, methods often rely on observing the variance across multiple predictive passes or distinct models. A seminal lightweight approach is Monte Carlo dropout, which interprets dropout at inference time as a Bayesian approximation of a Gaussian process, allowing for uncertainty estimation through stochastic forward passes \citep{dropout}. Alternatively, Deep Ensembles involve comparing the outputs of multiple models trained under different initialisations or data subsets \citep{lakshminarayanan2017simple}. In this context, ensemble prediction consistency serves as a direct proxy for reliability, where a high degree of disagreement between the models indicates that the input lies in an unconstrained region of the hypothesis space. When the ensemble fails to reach a consensus, it signals that the model's knowledge is limited, thereby identifying out-of-domain samples or regions of low data density.

\subsubsection{Sensitivity-Based Methods}
Robustness analysis evaluates model stability under perturbations of the input. Adversarial testing and Lipschitz-based assessments measure whether small, possibly physically plausible changes in the input induce disproportionately large changes in the output \citep{systematicliteraturereview, DEEL, EASA, controllingovergeneralizationeffectadversarial}. These techniques aim to detect fragile or unstable predictions. Nevertheless, robustness under local perturbations does not ensure correctness in globally shifted regimes.

\subsubsection{Training Data Attribution Methods}
These methods aim to enhance model explainability by identifying how individual training samples influence model behaviour. Influence functions constitute one such approach and can be employed as a diagnostic indicator by quantifying the effect of each training observation on a specific prediction. Rooted in robust statistics, the method approximates how the model parameters, and consequently the loss function, would change if a particular training sample were slightly upweighted \citep{hampel1974influence, Cook1977, cook1980characterizations, pmlr-v70-koh17a}. In situations where ground truth labels are unavailable, the classical formulation can be adapted to measure how upweighting a training point alters the predicted output itself and can be used as an effective error proxy \citep{nikkiIF}.

\subsubsection{Internal State Monitoring}
Activation pattern monitoring monitors the internal state of the neural network by establishing a baseline of neuron activation patterns during nominal training. During deployment, if a specific input triggers an unseen activation pathway, the system identifies that the model is processing information in a manner that was never formally validated \citep{runtimemonitoringneuronactivation, DEEL, APMMD}. This internal monitoring can catch points that may be within distribution but present an out-of-distribution activation pattern.

\subsubsection{Physics-Based Constraints}
In scientific applications, domain knowledge provides deterministic bounds on acceptable outputs. Physics-based filters impose constraints to ensure that predicted parameters remain physically plausible, such as preventing the planet radius from being larger than the star radius. These constraints are powerful for rejecting samples that present an impossible or unlikely physical behaviour \citep{ECSShandbook, safetyperspectiveonautonomousdriving}.

\subsubsection{Indicator Fusion}
\begin{table*}
\centering
\caption{Simulation components for the ADC19 and ADC21 datasets. Components are separated into those held fixed across both editions (shared simulation basis) and those that differ, constituting the sources of distribution shift between datasets. Stellar spots and limb darkening are present in both editions as part of the forward model, defining the core denoising challenge: recovering the transmission spectrum from spectroscopic light curves contaminated by both instrumental noise and stellar activity. LDE: limb darkening effect, modelled via Claret's 4-coefficient law \citep{claret2000}. MCS: Ariel Mission Candidate Sample.}
\label{tab:dataset_comparison}
\begin{tabular}{lcccc}
\toprule
& \multicolumn{2}{c}{\textbf{Fixed (both editions)}}
& \multicolumn{2}{c}{\textbf{Varies across editions}} \\
\cmidrule(lr){2-3}\cmidrule(lr){4-5}
\textbf{Dataset}
    & \textbf{Stellar \& orbital}
    & \textbf{Observation model}
    & \textbf{Noise model}
    & \textbf{Atmospheric composition} \\
\midrule
ADC19
    & MCS (same)
    & Spots $+$ LDE (same)
    & Photon noise (ArielRad-like)
    & H$_2$O, CH$_4$ \\
ADC21
    & MCS (same)
    & Spots $+$ LDE (same)
    & Photon $+$ $1/f$ $+$ persistence
    & H$_2$O, CH$_4$, CO$_2$ \\
\bottomrule
\end{tabular}
\end{table*}

No single monitoring mechanism captures all failure modes. Consequently, safety cage architectures should employ multi-layered monitoring and indicator fusion. Fusion strategies may involve sequential gating, weighted aggregation into a continuous safety score, or learned meta-classifiers that approximate a decision boundary between reliable and unreliable predictions. Regardless of the specific method, the objective of indicator fusion is to collapse the multi-dimensional safety evaluation space into a singular action that ensures the operational integrity of the system during operations.

A central challenge in fusion design is balancing interpretability and robustness. Overly complex fusion schemes may obscure decision logic, while simplistic thresholds may fail under distributional shift. Furthermore, it is essential to perform a systematic evaluation of fusion strategies in relation to domain-specific metrics. However, finding the ideal metric for a domain is also a challenge in itself.

\subsection{Scientific Gap and Research Questions}
Although uncertainty quantification, out-of-domain detection, and robustness analysis have been widely studied \citep{safetyandtrustworthiness, gawlikowski2021survey, goodfellow2015adversarial, chandola2009anomaly, yang2024generalized}, their effectiveness in scientific inference pipelines without access to ground truth remains inadequately characterised \citep{corbiere2019failure, carleo2019machine}. Most existing evaluations assume labelled validation data, whereas operational scientific missions must infer reliability solely from runtime observables.

A primary unresolved challenge is failure detection without reference labels: how can model errors be identified when no ground truth is available? Relatedly, it remains unclear whether safety mechanisms can meaningfully bound a model’s effective operational domain. Furthermore, the behaviour of safety indicators under covariate shift and concept drift has not been rigorously compared within controlled scientific settings. Finally, beyond detecting clear out-of-domain failures, it remains uncertain whether runtime monitoring can identify regions beyond the training data where the model remains valid or regions within the training data that have been wrongly constrained.

These gaps motivate a structured empirical investigation of safety cage architectures under controlled in-domain and out-of-domain settings in exoplanet spectrum reconstruction.

\section{Data}
ESA's Ariel mission serves as an operational use case to investigate the implementation of runtime safety cages within a high-stakes scientific pipeline \citep{tinetti2018chemical, tinetti2021arielenablingplanetaryscience}. In this case, the ML task consists of extracting transmission spectra from time-series light curves. This task requires high precision, as atmospheric absorption signatures introduce atmospheric transit depth variations as faint as a few dozen ppms relative to the total stellar flux.

\begin{figure*}
    \centering
    \includegraphics[width=0.8\linewidth]{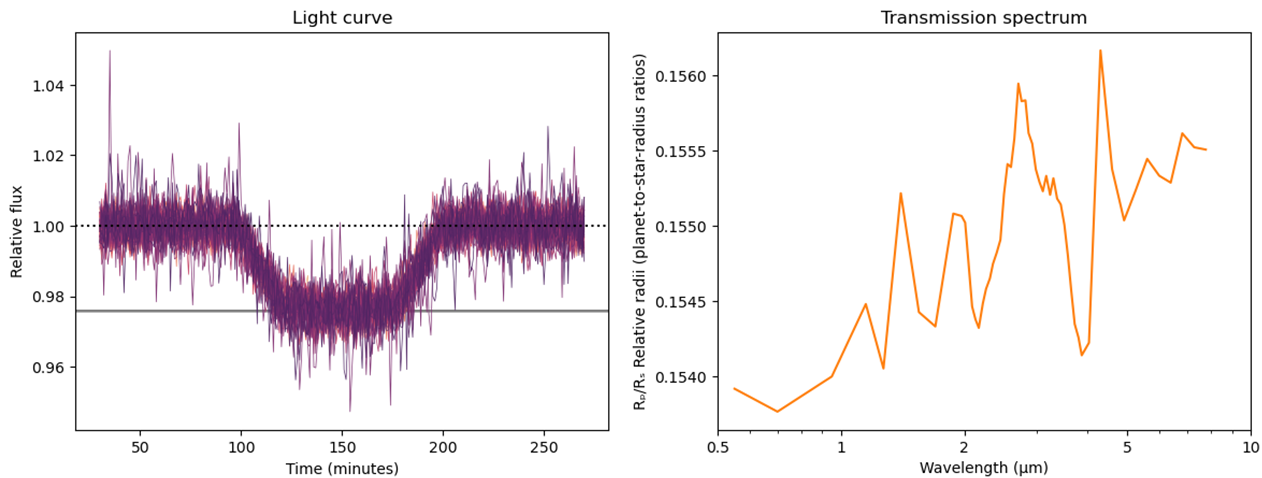}
    \caption{Observed transit light curves (left) and corresponding transmission spectrum (right) for an exoplanet. There are 55 light curves that correspond to 55 wavelength channels that translate to one transmission spectrum. }
    \label{fig:lightcurve_transmission}
\end{figure*}

The datasets used for this research were derived from the 2019 and 2021 editions of the Ariel Data Challenge \citep{nikolaou2023lessons, yip_2025_15050868}. These competitions, organised by mission scientists and engineers, invited the global research community to develop innovative solutions for complex data analysis tasks associated with the mission. The 2019 and 2021 challenges specifically addressed the task of extracting planetary spectra from noisy time-series light curve data contaminated by stellar spots. These datasets use ExoSim \citep{Sarkar2021} to produce realistic noise profiles mimicking real observations, including stellar activity and instrumental systematics, simulating the SNR constraints inherent in the mission's spectroscopic observations. A description of the simulation of the datasets can be found in \autoref{tab:dataset_comparison}, and an example of how this data looks for an exoplanet can be found in \autoref{fig:lightcurve_transmission}.

The noise profiles used to simulate the ADC19 and ADC21 datasets differ significantly from each other. In ADC19, the simulation considered only stellar noise, assuming an ideal, noise-free instrument. In contrast, ADC21 incorporated a more realistic setup by convolving the signal with non-linear, time-dependent instrument response models, introducing detector persistence and non-Gaussian red noise \citep{Sarkar2021, safetycages}. For each exoplanet, light curves were generated under 10 different stellar spot instances, each under 10 different photon noise instances, resulting in 100 simulations per exoplanet. This was done for many different exoplanets, with 851 being present in both datasets, representing the samples used in this analysis. The simulations include 55 wavelength channels, each with 300 time points, corresponding to 5 hours of measurements from Ariel's sensors.

From an ML perspective, these differences across datasets enable the study of model performance and generalisation under conditions of data shift. Covariate shift is observed when the input feature distributions differ between the simplified ADC19 and the instrument-contaminated ADC21, for example, due to the introduction of non-linear detector artefacts. Simultaneously, concept drift occurs when the fundamental relationship between the light curve and the target spectrum is altered by time-dependent response models, as happened in a portion of the samples of ADC21 that contained persistence noise, a non-linear time-dependent change in detector's sensitivity.

\section{Experimental Setup}
The experimental setup is structured to evaluate the safety cage's ability to monitor model integrity in real-time. The framework combines a predictive model with a set of diagnostic indicators designed to detect conditions under which predictions may become unreliable.

\subsection{Preprocessing Pipeline}
The process begins with pre-processing, where the raw time-series data from ADC19 and ADC21 are aggregated following a sequence of steps designed to increase the SNR. First, the 10 photon noise instances are treated as repeated transits of the same planet and are aggregated to form a more stable light curve. Because the transits are centred around the 150-minute mark, the exact midpoint of the time series, a symmetry-based aggregation is applied. Observations are folded over this midpoint, combining points that are temporally symmetric to the left and right of the transit centre. This was done because, for an idealised transit without stellar variability or instrumental systematics, the light curve is expected to be approximately symmetric around the transit midpoint. Averaging corresponding points on either side of the midpoint can, therefore, reduce the influence of local asymmetric perturbations, such as spot-crossing events, while preserving the dominant transit shape. Furthermore, a temporal window is applied to the aggregation process, incorporating $n$ points to the left and right of a given timestamp. For this research, a window radius of nine minutes was selected, and the values were combined using a harmonic mean to reduce the influence of extreme outliers and high-frequency fluctuations. This resulted in each point being an aggregation of $10 \cdot 2 \cdot (2 \cdot 9 + 1) = 380$ points. Following these aggregation steps, a reduced feature grid is obtained by selecting five key sampling points that begin at 20\% of the transit duration and extend evenly to the midpoint, effectively compressing the high-dimensional time series into a set of representative features that capture the dynamics of ingress and mid-transit. This approach follows the idea presented by \citep{safetycages}.

\subsection{Feature Representation}
In addition to these structural representations of the light curve, several auxiliary features are integrated into the feature store to provide the model with a comprehensive context for each observation. This set includes astrophysical parameters sourced directly from the ADC19 and ADC21 metadata, the version of the MCS used in the data generation process, and various parameters derived from these primary sources \citep{edwards2019updated}. Specifically, the feature set incorporates stellar and planetary characteristics, such as stellar density, temperature, log gravity, radius, mass, k-mag, distance, luminosity, and incident flux, alongside the geometric and orbital properties of the system, including transit duration, eccentricity, impact parameter, semi-major axis, inclination, and orbital period.

\subsection{Predictive Models}
The modelling approach comprises two distinct philosophies to assess their efficacy for spectral reconstruction. The first is Ridge Regression, which serves as a linear baseline that employs an $L_2$ regularisation to manage the multicollinearity inherent in highly correlated wavelength channels and astrophysical features \citep{mcdonald2009ridge}. To capture the complex, non-linear relationships within the data, the Extreme Learning Machine (ELM) was also implemented \citep{HUANG2006489, tara_elm, huang2015trends, wang2022review}. This approach utilises a single hidden-layer feedforward neural network where the input weights are randomly assigned and remain fixed, and the output weights are analytically determined using Ridge Regression to solve the least-squares problem, allowing for extremely rapid training. The ridge regression model was initialised using the default parameters of the \texttt{scikit-learn} implementation \citep{scikit-learn}. The ELM architecture consists of a single hidden layer containing 5000 neurons with a sigmoid activation function. The output layer weights are estimated through ridge regression with a regularisation parameter of $\alpha = 1.0$. In addition, the ELM incorporates skip connections from the input to the output layer. The input-to-hidden layer connection weights are initialised using the strategy described in \cite{elm_input_to_hidden}.

Prior to training, the targets were transformed using a power encoding defined as $y_{\text{enc}} = y^{1/4}$. After prediction, the outputs were mapped back to the original scale using the inverse transformation. Empirically, this encoding produced the most stable training behaviour and the lowest prediction errors among the tested transformations, and was therefore adopted in the final experimental setup.

The compatibility with safety cage monitoring was only evaluated on the ELM. Ridge regression is nevertheless included, as a safety cage architecture was previously developed and validated using this modelling approach by \cite{safetycages} on the same dataset, and is now extended to a more complex model. Where possible, results are compared at a high level with those obtained for ridge regression.

\subsection{Evaluation Metrics}
To provide a comprehensive view of the model's accuracy and its relationship with the safety cage's indicators, two evaluation metrics are used to capture both scale- and shape-based performance. The Root Mean Squared Percentage Error (RMSPE) is used as a scale-based metric to account for relative error across varying transit depths, ensuring the signal magnitude is correctly scaled. Complementing this, the Spectral Angle Mapper (SAM) serves as a shape-based metric by treating the predicted and ground truth spectra as vectors in a multi-dimensional space and measuring their geometric similarity independent of absolute brightness \citep{kruse1993spectral}.

During the experimental phase, the availability of ground truth for simulated data enables direct calculation of performance metrics. However, during the operational phase, ground truth is unavailable, making it impossible to obtain a direct score of model performance. For this reason, it is critical to identify safety indicators that strongly correlate with the evaluation metrics. Furthermore, the metrics must be selected carefully to ensure they accurately reflect the prediction quality and the successful extraction of atmospheric information. By establishing that a metric accurately reflects scientific validity and subsequently identifying indicators that correlate with it, the safety cage can reliably distinguish between ``good'' and ``bad'' predictions in real time, based solely on the observed behaviour of the safety indicators.

\begin{figure}
    \centering
    \includegraphics[width=\linewidth]{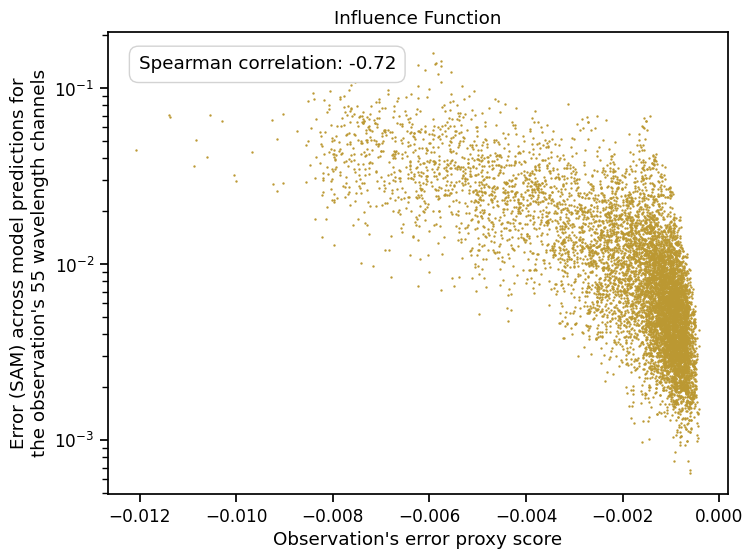}
    \caption{Validation of the influence function error proxy. Relationship between the safety indicator scores (derived via cross-validation on the ADC21 dataset) and the true predictive performance of the ELM model. The degradation in model performance, shown by higher SAM errors, as safety scores decrease, demonstrates the framework's effectiveness in ranking samples by risk.}
    \label{fig:scatter_if}
\end{figure}

\subsection{Experimental Design}
To evaluate the safety cage's efficacy, a 5-fold cross-validation (CV) approach is used to generate safety scores for the out-of-fold samples. This structured validation ensures that every data point in the training set is treated as an unseen sample at least once, providing a robust distribution of runtime safety indicators. As a result of the preprocessing pipeline, each exoplanet is represented by 10 samples, one per stellar spot configuration. To avoid data leakage and contamination, the CV strategy splits the samples at the exoplanet level. By correlating these runtime safety scores with true performance metrics, it is possible to evaluate whether an indicator's degradation results in a corresponding drop in prediction quality. This relationship is visualised in \autoref{fig:scatter_if}, which displays the error proxy and true error for each sample alongside its Spearman rank correlation coefficient ($\rho$). The Spearman correlation is a non-parametric measure that quantifies the strength and direction of a monotonic relationship between two variables \citep{spearman1904proof}. Unlike the standard Pearson correlation, which assumes a strictly linear relationship, the Spearman coefficient evaluates how well the relationship between the safety score and the performance metric can be described using an arbitrary monotonic function. A strong negative correlation indicates that as the safety score decreases, the performance error for that sample increases, confirming the indicator's utility.

\begin{figure*}
    \centering
    \includegraphics[width=\linewidth]{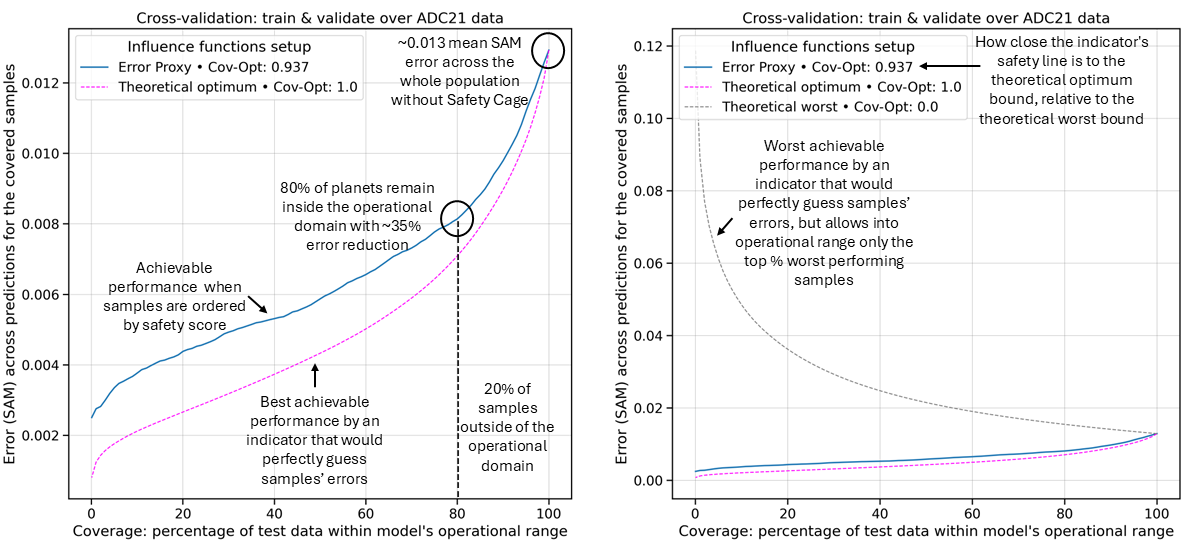}
    \caption{Operational coverage–error curves for the influence function safety indicator. Both panels show the framework performance for in-domain validation on the ADC21 dataset, evaluated using the SAM error metric. The left panel illustrates the indicator's ability to reject higher-error samples as coverage is reduced progressively. The right panel includes the theoretical worst-case baseline, in which the highest-error samples are accepted first, thereby highlighting the gain achieved by the safety cage architecture.}
    \label{fig:explaine_cov_err}
\end{figure*}

Models are first cross-validated within the same dataset to assess whether safety indicators correlate with predictive error under nominal conditions. Afterwards, cross-validation evaluates models on the other dataset to assess whether safety indicators degrade under distributional shifts and whether they can detect these shifts. By ranking predictions by safety score, high-risk samples are progressively filtered out so that aggregated performance metrics can be recomputed on the remaining subset. This approach allows validation metrics to be evaluated across multiple coverage thresholds, demonstrating how the model performs when restricted to specific top percentages of the safest data. This analysis quantifies the operational benefit of safety-driven rejection strategies in reducing performance error and results in the generation of the coverage-error plots used during analysis as introduced in \cite{selective_classification}. A schematic guide for the interpretation of these plots is presented in \autoref{fig:explaine_cov_err}.

From the coverage-error curve obtained by sequentially adding increasing fractions of high-risk samples, a metric based on the area under the curve (AUC) is computed \citep{hanley1982auc}. This metric is referred to as \textit{Cov-Opt} (``Coverage Optimality''), and a value of 1 corresponds to the theoretical optimum, while a value of 0 represents the theoretical worst case. The optimal case is obtained when samples are ordered according to their true error values, such that the smallest errors are identified first. Conversely, the worst case corresponds to the inverted ordering, where samples with the highest errors are selected first. By comparing the area under the indicator-based curve with these two theoretical bounds, the metric quantifies how close the indicator comes to the ideal scenario in which the safety signal is perfectly correlated with the prediction error. For evaluation metrics where lower values correspond to better performance, the Cov-Opt score is defined as

\begin{equation}
\label{eq:cov-opt}
\text{Cov-Opt} = 1 - \frac{AUC_{\text{Indicator}} - AUC_{\text{Best}}}{AUC_{\text{Worst}} - AUC_{\text{Best}}}.
\end{equation}

This formulation normalises the indicator's performance between the theoretical best and worst orderings, thereby providing a direct measure of how effectively it prioritises low-error predictions, also allowing comparison between different models. Based on the formulation, this metric corresponds to the Area Between Curves (ABC). Indicators that exhibit strong negative correlation with the evaluation metric and favourable coverage-risk trade-offs are considered viable candidates for operational safety cage deployment.

To analyse how the different indicators relate to the problem, the strongest-performing indicator within each failure identification method is first identified. Subsequently, the indicators are compared, and the potential benefits of combining them via indicator fusion are assessed.

\section{Results}
The baseline performance of the two modelling architectures was calculated, and the results are presented in \autoref{tab:performance}. These values demonstrate a clear progression in predictive accuracy across the two evaluated architectures from simple to more complex. The Ridge Regression model serves as a baseline, exhibiting the highest error margins across the two metrics for the two evaluation setups. In contrast, the ELM provides a substantial performance improvement, with the highest reduction being for RMSPE 21 error (RMSPE of the entire model when in cross-validation with training and testing on ADC21) by approximately 60\% compared to the linear baseline.

\begin{table}
    \centering
    \caption{Performance comparison of Ridge Regression and ELM architectures for RMSPE and SAM in the cross-validation setting when training on the ADC21 dataset and testing on both the ADC21 and the ADC19 datasets. These evaluations correspond to a safety cage with 100\% coverage (no sample filtering).}
    \begin{tabular}{c|c|c|c|c}
    \hline
      Indicator & RMSPE 21 & RMSPE 19 & SAM 21 & SAM 19\\ \hline
     Ridge Regression  &  0.136166 & 0.131880 &  0.017489 &  0.022598\\
     ELM  & 0.054703  & 0.069541 &  0.012931  & 0.017966 \\ \hline
    \end{tabular}
    \label{tab:performance}
\end{table}

After evaluating the predictive performance of the base models, the next step is to assess the effectiveness of the proposed safety indicators in identifying potentially unreliable predictions and enabling the operation of the safety cage. This was done only for the ELM due to its increased predictive performance.

\subsection{Safety Indicators}
The runtime monitoring architecture relies on six distinct families of safety indicators, each engineered to capture independent and complementary failure modes of the predictive models. Rather than serving as an exhaustive comparison of every candidate metric, this section outlines the final configuration chosen for each family, the systematic derivation, evaluation, and selection of which are detailed in \autoref{App:safetyInd} of the Appendix. A summary of these optimal selections, along with how to interpret the indicators, is provided in \autoref{tab:safety_indicators}. These indicators are formulated to rank the reliability of individual model predictions during deployment. Operationally, a higher safety score signifies greater confidence that an input resides within the model's competence range.

\begin{table*}
\centering
\small
\caption{Comprehensive summary of the selected safety cage monitoring framework. For each of the six indicator families, the table outlines the chosen metric, its underlying computational mechanism, and the operational interpretation of its safety score.}
\label{tab:safety_indicators}
\begin{tabular}{l p{2.2cm} p{4.5cm} p{5.5cm}}
\toprule
\textbf{Indicator} & \textbf{Best Variant} & \textbf{Underlying Mechanism} & \textbf{Operational Meaning of High Score} \\ 
\midrule
Out-of-Distribution & Isolation Forest on $X_{\text{shap}}$ & Isolation Forest applied to 30-D PCA-reduced feature attributions (SHAP values). & Feature importance logic aligns with safe, in-domain training distributions. \\
\noalign{\smallskip}
Uncertainty Quantification & $\text{Width}$ & Conformalised Quantile Regression via Multi-Output LightGBM on error PCA space. & High model certainty shown by narrow predictive intervals across spectral channels. \\
\noalign{\smallskip}
Ensemble Consistency & $\text{Relative Uncertainty}$ & Relative ensemble variance benchmarked against typical training phase disagreement. & High model consensus with live relative variance not exceeding historical training baselines. \\
\noalign{\smallskip}
Adversarial Attack & $\text{Risk}$ & A perturbation cloud based on MCS error profiles to evaluate relative prediction drift and variance. & High localised model robustness with the model having low sensitivity to small input perturbations. \\
\noalign{\smallskip}

Influence Function & $\text{Error Proxy}$ & Variance estimation of training set influence via the Infinitesimal Jackknife. & Low reliance on poorly modelled training data. \\
\noalign{\smallskip}
Activation Pattern Monitoring & $\text{Mahalanobis } Z_s$ & Mahalanobis distance calculation on 50-D PCA-reduced standardised hidden layer activations. & Internal neural network states match normal, expected activation pattern. \\
\bottomrule
\end{tabular}
\end{table*}

For OOD detection, the final selection is $X_{\text{shap}}$, which executes anomaly detection via an Isolation Forest model. To manage the high dimensionality of the operational dataset, comprising 291 input features, these are projected into a manageable 30-dimensional space using Principal Component Analysis (PCA) \citep{pearson1901liii, safetycages}. Rather than monitoring raw inputs, the indicator operates within the explanation space by utilizing SHAP values derived from a Ridge-based linear explainer \citep{lundberg2017unified}. By encoding inputs as the median SHAP values across the 55 spectral dimensions, $X_{\text{shap}}$ monitors whether the underlying feature attribution patterns of a live prediction remain consistent with the logic established during training. This configuration allows the anomaly detection framework to evaluate the mapping mechanism from input features to output targets, effectively monitoring the interaction between both spaces simultaneously.

To quantify predictive uncertainty, the UQ framework uses the Width indicator, which combines dimensional reduction with conformalised quantile regression. A Multi-Output LightGBM Regressor \citep{ke2017lightgbm} is trained to estimate the 5th and 95th percentiles of error terms previously projected into a lower-dimensional PCA space, thus accounting for the physical correlation between adjacent wavelength channels. Next, the lower and upper prediction bounds are established. These are calculated by combining the quantiles with conformity scores from a calibration set, which is generated via nested cross-validation (where an inner cross-validation loop runs on each training set). The final safety score is defined as the negative mean of the relative conformal interval width across all wavelengths, meaning that wider intervals directly scale down the safety score to reflect increased model uncertainty.

For Ensemble Consistency (EC), the Relative Uncertainty metric was selected. The framework first divides the ensemble standard deviation by the ensemble mean, to ensure the value is within the magnitude of the signal. Afterwards, the negative of this value is divided by the baseline standard deviation calculated across the training distribution for each wavelength channel, allowing for contextualisation. Whenever live inference disagreement surpasses the typical levels seen during training, the safety score reduces, flagging the input as a scenario the ensemble cannot predict with certainty.

Robustness under input perturbations is monitored via the Risk indicator, which evaluates model stability when subjected to Adversarial Attacks (AA). By perturbing the test inputs based on characteristic errors derived from the MCS, a cloud of $N$ physically plausible samples is generated around each test point. The Risk indicator combines both the magnitude of the resulting prediction shift and the overall standard deviation of the model's response across the 55 spectral channels for all $N$ cloud points. A high risk profile indicates that the model output is overly sensitive to small input fluctuations, signifying a lack of localised robustness or the presence of a steep error gradient around that specific input vector. A key limitation of this setup is its reliance on synthetic data for the light curves, transmission spectra, and physical parameters. Because the simulator creates a perfect mathematical relationship between these variables, it is difficult to test how the system reacts to slight mismatches in the input space, as these inconsistencies never occur within the training data. In real-world operations, however, physical parameters are estimated independently and always carry errors, meaning this perfect alignment is broken. In a deployment setting, it is important to ensure that a small, realistic perturbation in the inputs does not trigger a large jump in the model's prediction.

The Error Proxy indicator from Influence Functions (IF) provides a statistical estimate of the potential prediction error by leveraging training set influence via the Infinitesimal Jackknife estimator variance \citep{jaeckel1972infinitesimal, Jacknifeplusinfluence, swissIJ, nikkiIF}. By calculating the root mean of the squared influences across all wavelengths and normalising by the model's prediction, the negative of this value acts as a reliable confidence proxy. High values indicate that a prediction is heavily influenced by wrongly modelled training samples.

Lastly, Activation Pattern Monitoring (APM) via the $\text{Mahalanobis }Z_s$ indicator assesses the internal state of the neural network by calculating the Mahalanobis distance between live hidden-layer activation patterns and the baseline distribution established during training \citep{mahalanobis1936distance}. Both the live and baseline activations are standardised using the mean and variance characteristics of the training dataset. To maintain computational efficiency, PCA is first applied to reduce these hidden-layer activations to the top 50 principal components. A large Mahalanobis distance indicates that the model is processing data through an unfamiliar internal activation pattern.

\subsection{Indicator Fusion}
\begin{figure*}
    \centering
    \includegraphics[width=0.8\linewidth]{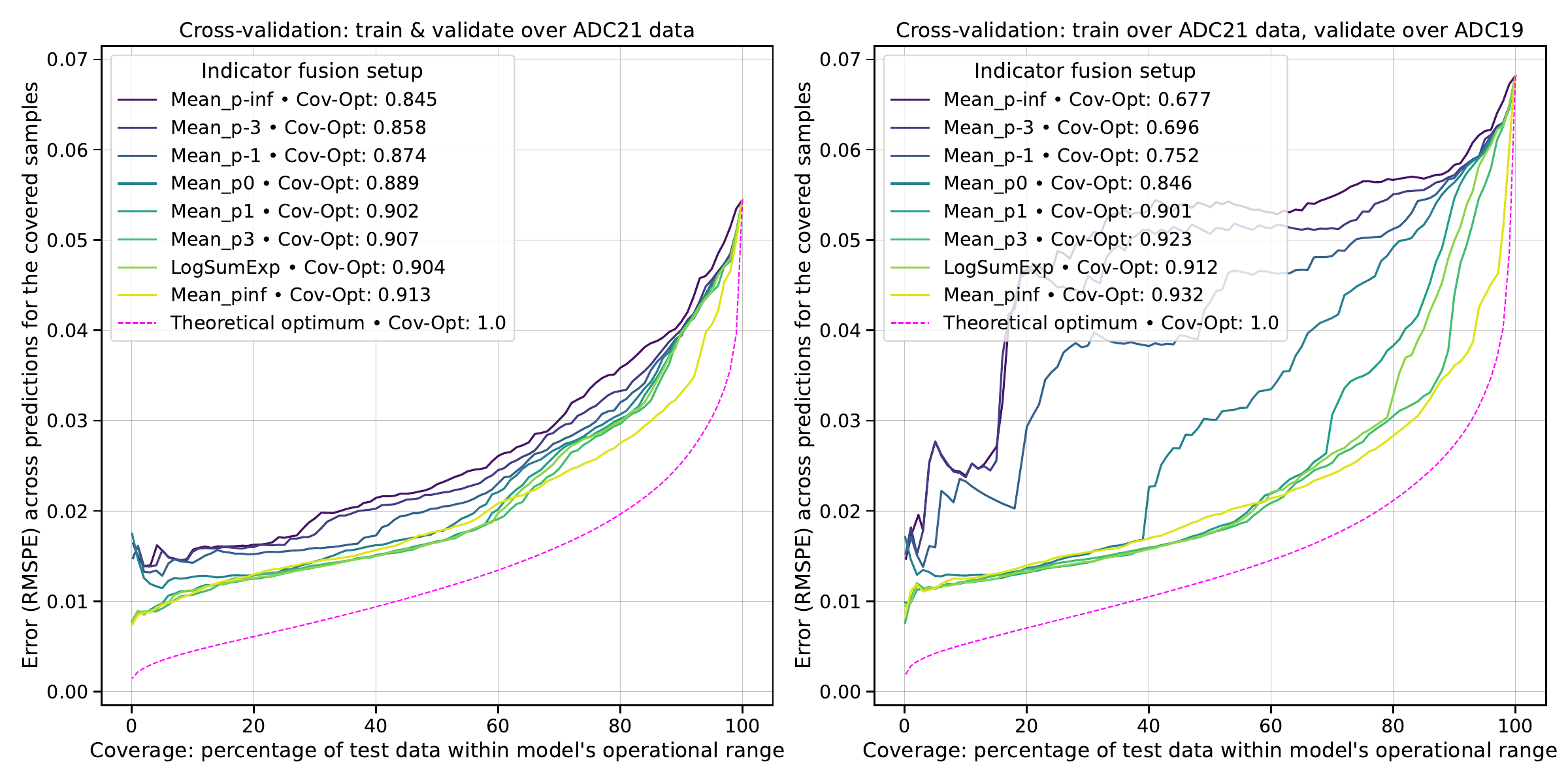}
    \caption{Illustration of RMSPE evolution as samples are admitted to the model's operational range, based on fused safety scores generated through various generalised mean strategies and LogSumExp, comparing internal (ADC21) and cross-domain (ADC19) validation.}
    \label{fig:FusionRMSPE}
\end{figure*}

\begin{figure*}
    \centering
    \includegraphics[width=0.8\linewidth]{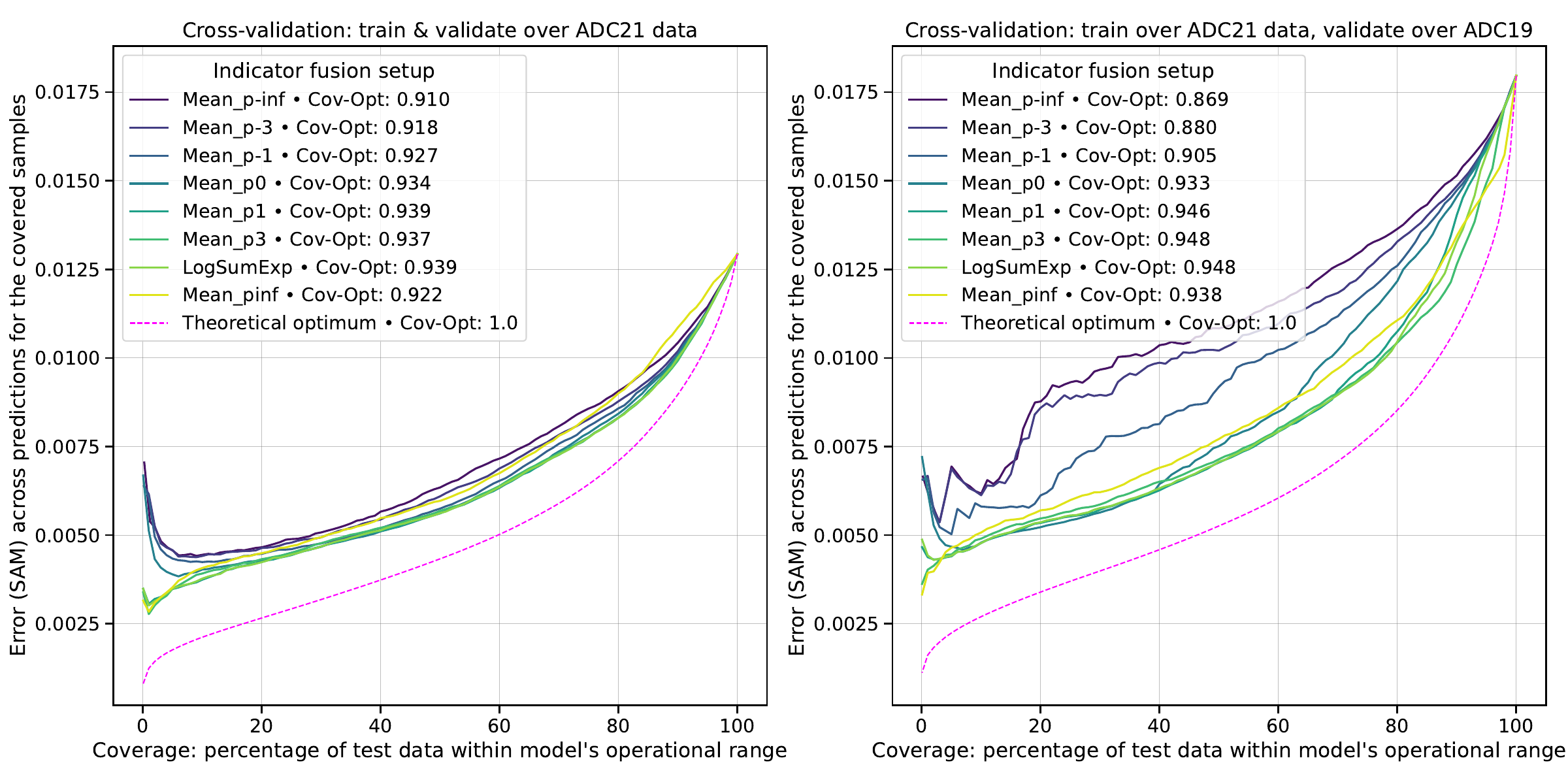}
    \caption{Illustration of SAM evolution as samples are admitted to the model's operational range, based on fused safety scores generated through various generalised mean strategies and LogSumExp, comparing internal (ADC21) and cross-domain (ADC19) validation.}
    \label{fig:FusionSAM}
\end{figure*}

Following the identification of the top-performing safety indicators from each distinct family in the previous section, specifically, Isolation Forest on $X_{\text{shap}}$, Width, Relative Uncertainty, Risk, Error Proxy, and Mahalanobis $Z_s$, the analysis now progresses to the indicator fusion framework. To leverage the complementary strengths of these individual metrics, several aggregation techniques based on generalised means and the LogSumExp function are evaluated. The primary objective is to determine the most effective mathematical formulation for combining the different safety indicators into a single, robust safety score.

Before applying any aggregation method, the raw safety scores are first transformed to a common risk scale using a quantile-based normalisation derived from the training distribution. For each indicator, the empirical cumulative distribution function (ECDF) is estimated using the training scores. Each score, both from the training and test sets, is then mapped to its corresponding empirical quantile with respect to this training distribution. To allow for an interpretation of risk across indicators, the quantile is mapped so that higher safety scores correspond to low risk, while lower safety scores map to higher risk. This procedure effectively expresses each indicator as the relative position of a sample within the training distribution, producing values on a comparable $[0,1]$ risk scale while avoiding assumptions about the underlying score distributions. Although this process transforms the safety scores into risk scores, coverage remains defined by the inclusion of samples with the lowest risk and highest safety. Using the training distribution as the reference also prevents information leakage from the test set and ensures that the risk calibration reflects only the behaviour observed during model training.

The first aggregation technique corresponds to LogSumExp. This method serves as a smooth approximation of the maximum function. By calculating the logarithm of the sum of the exponentials of all individual risk scores, the system prioritises the highest risk indicators while allowing others to contribute to the final value \citep{LSE}. The next technique is the Generalised Mean, which offers a flexible aggregation framework controlled by the power parameter $p$ \citep{hardy1952inequalities}. By adjusting $p$, the fusion sensitivity shifts toward different safety priorities. A parameter of $p = +\infty$ represents the most conservative safety cage, where the final score equals the maximum of all individual risk indicators and translates to: if any single check fails, the entire prediction is flagged as unsafe. Conversely, $p = -\infty$ represents the most optimistic approach, following the minimum indicator value. Intermediate values provide nuanced weighting: $p = -3$ and $p = -1$ (Harmonic Mean) impose strong biases in favour of low risk scores, ensuring that good performance in even one safety area significantly increases the overall reliability rating. A value of $p = 0$ corresponds to the geometric mean, which balances the contributions of all indicators multiplicatively and penalises situations where several indicators simultaneously report elevated risk. A value of  $p = 1$ corresponds to the arithmetic mean, where all perspectives contribute equally, while $p = 3$ shifts the aggregation bias toward higher risk indicators without completely ignoring lower and safer ones. As $p$ increases, the aggregation progressively emphasises the largest risk values, approaching the behaviour of a maximum operator. Conversely, decreasing $p$ increases the influence of lower values, favouring optimistic interpretations of the safety indicators.

\begin{figure*}
    \centering
    \includegraphics[width=0.8\linewidth]{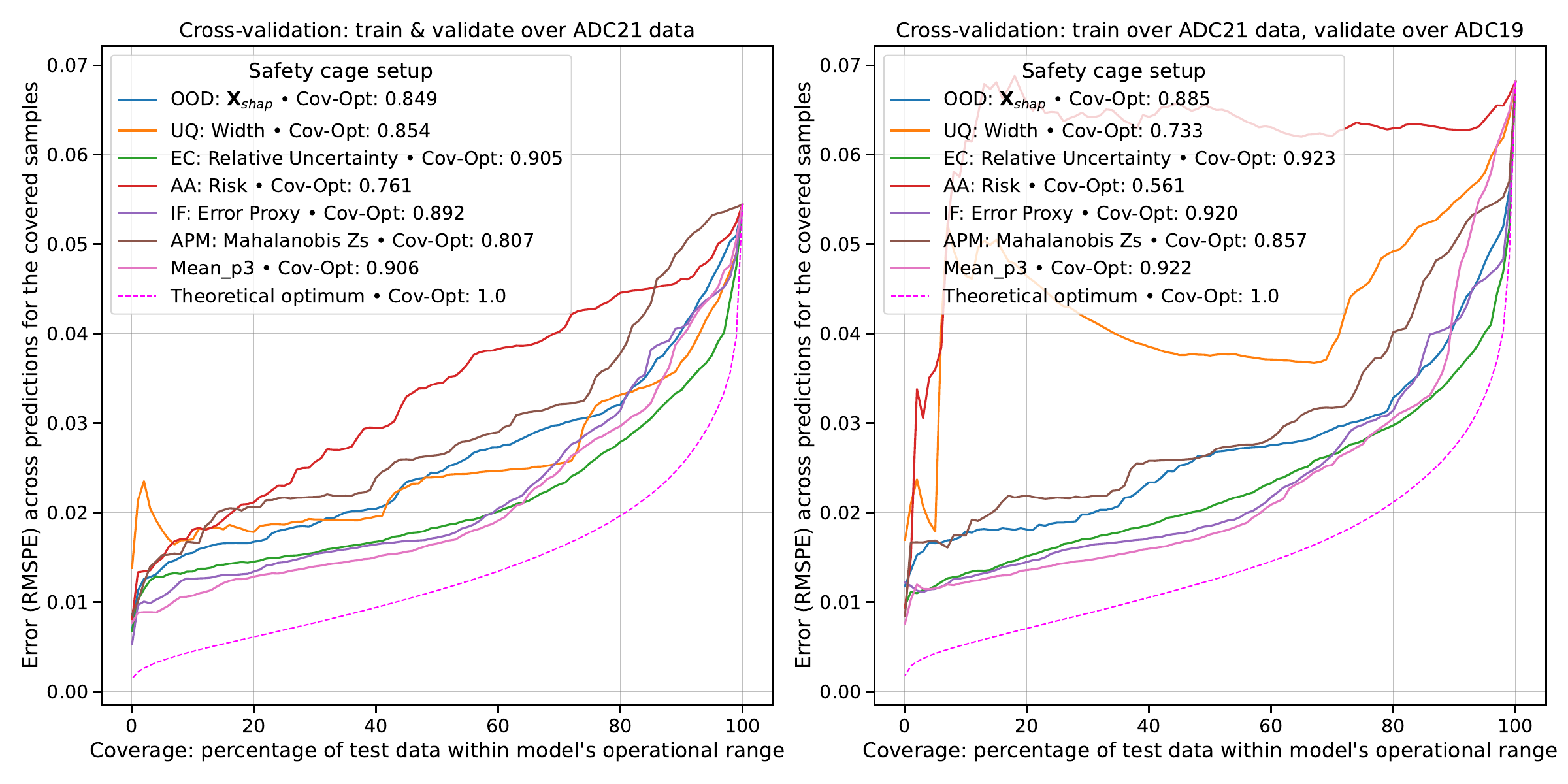}
    \caption{Illustration of RMSPE evolution as samples are admitted to the model's operational range based on the best safety indicators for each setup, as well as the best fusion strategy. The left panel shows internal validation on ADC21, while the right panel demonstrates cross-domain validation on ADC19.}
    \label{fig:FinalRMSPE}
\end{figure*}

\begin{figure*}
    \centering
    \includegraphics[width=0.8\linewidth]{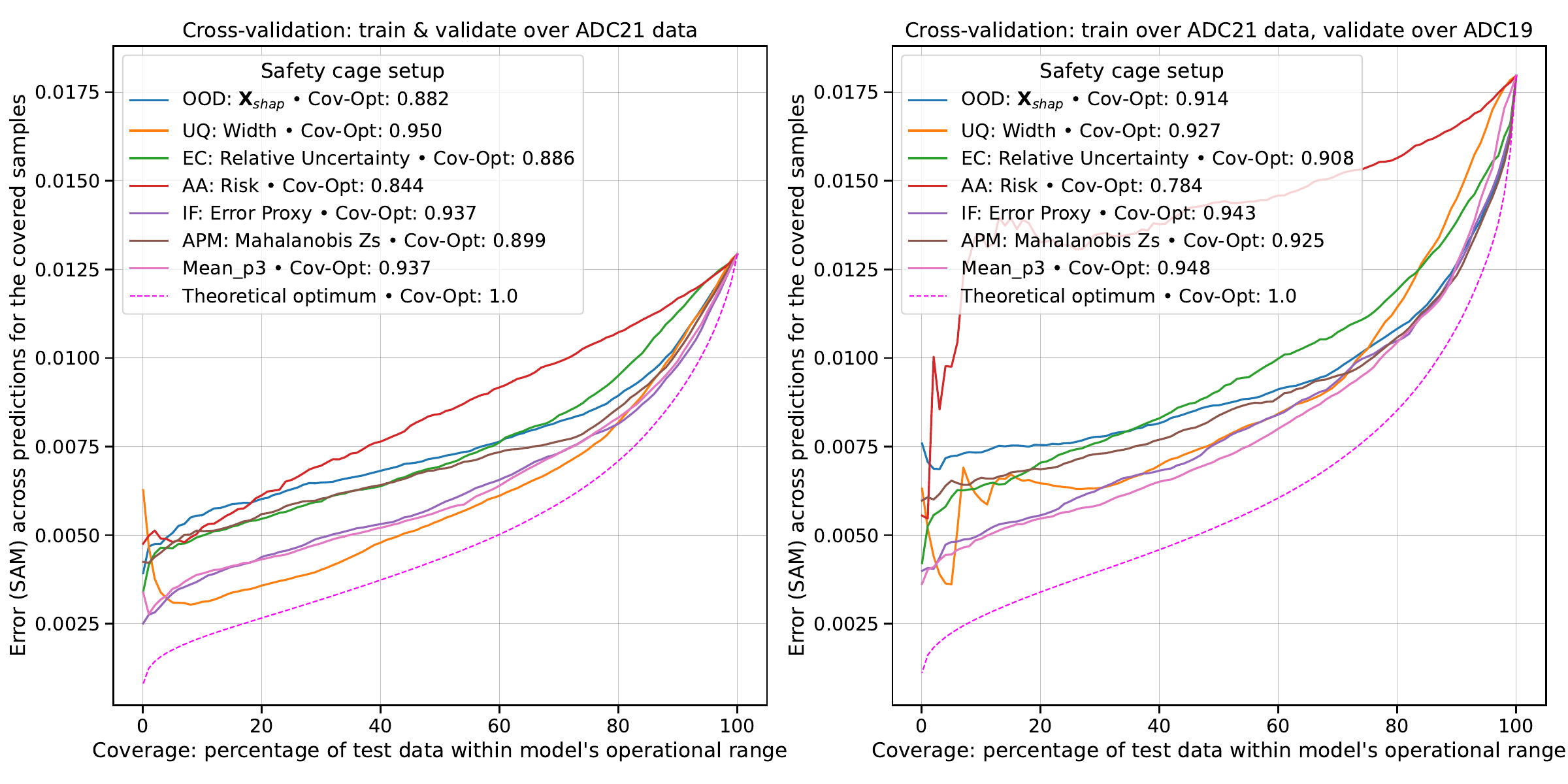}
    \caption{Illustration of SAM evolution as samples are admitted to the model's operational range based on the best safety indicators for each setup, as well as the best fusion strategy. The left panel shows internal validation on ADC21, while the right panel demonstrates cross-domain validation on ADC19.}
    \label{fig:FinalSAM}
\end{figure*}

\begin{table}
    \centering
    \caption{Spearman correlation coefficients between prediction error metrics and fused safety scores calculated using various generalised mean strategies and LogSumExp across the ADC21 and ADC19 datasets.}
    \begin{tabular}{c|c|c|c|c}
    \hline
      Indicator   & RMSPE 21 & RMSPE 19 & SAM 21 & SAM 19\\ \hline
     Mean\_p-inf  & -0.605 &  -0.597 & -0.672& -0.638\\
     Mean\_p-3  & -0.630 & -0.625 & 	-0.694 & 	-0.661\\
     Mean\_p-1  & -0.660 & -0.664 & -0.718 & -0.691\\
     Mean\_p0  & -0.687 & -0.699 &  -0.735 & -0.714\\
     Mean\_p1  & -0.703 & -0.716 &  -0.738 & -0.716	\\
     Mean\_p3  & -0.707 & -0.714	 & -0.728 & -0.700\\
     LogSumExp & -0.706  &  -0.718 & -0.736 & -0.714\\
     Mean\_pinf  & -0.673 & -0.667  &  -0.669 & 	-0.635\\
      \hline
    \end{tabular}
    \label{tab:FusionCorr}
\end{table}

The Spearman correlation coefficients presented in \autoref{tab:FusionCorr} show that the choice of the power parameter $p$ in the generalised mean has a significant influence on the performance of the indicator fusion. As  $p$ increases from negative values toward positive values, the correlation between the fused safety score and the prediction error generally improves. This behaviour is consistent with the safety interpretation of the indicators: if any individual safety check signals a potential issue, the fused score should reflect this increased risk. The results indicate that aggregation methods with moderate emphasis on higher risk values perform best. In particular, LogSumExp and the generalised mean with $p = 3$ and $p = 1$ achieve overall the strongest correlations across both datasets and both error metrics. In contrast, strongly optimistic aggregation strategies that emphasise low risk indicators (negative $p$) show consistently weaker correlations.

The coverage-error curves in \autoref{fig:FusionRMSPE} and \autoref{fig:FusionSAM} provide further insight into the operational behaviour of the fusion techniques. For RMSPE, the highest Cov-Opt values are obtained by Mean\_pinf, Mean\_p3, and LogSumExp. When analysing SAM, the best performing techniques are LogSumExp, Mean\_p3, and Mean\_p1. Considering the overlap between the best-performing methods across both metrics, as well as the magnitude of the achieved scores, Mean\_p3 and LogSumExp emerge as the best aggregation strategies.

Based on these results, Mean\_p3 is selected as the fusion strategy. This choice is supported by both its strong statistical correlation with the prediction errors and its behaviour in the coverage-error analysis. Furthermore, Mean\_p3 demonstrates stable behaviour across the transition from ADC21 to ADC19, indicating robustness to domain shift. When looking at indicator fusion using Mean\_p3, with only a 20\% reduction in coverage, there is a 45\% reduction in RMSPE for ADC21 and more than a 65\% reduction for ADC19. For SAM, the reductions reach approximately 35\% for ADC21 and 40\% for ADC19. This demonstrates that the fusion of safety indicators allows the safety cage to effectively identify and filter high-risk predictions while preserving the majority of the data coverage.

\begin{figure*}
    \centering
    \includegraphics[width=0.75\linewidth]{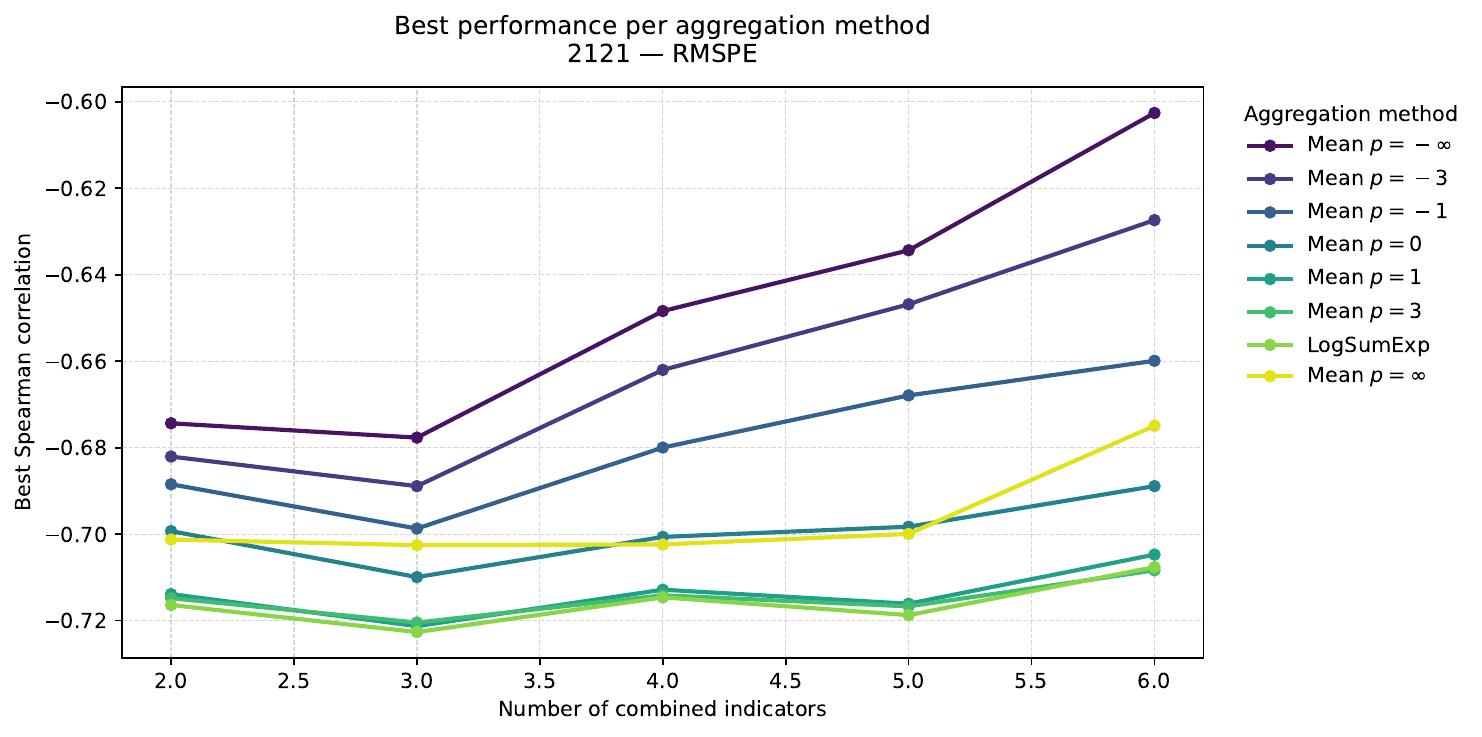}
    \caption{Best Spearman correlation per aggregation method as a function of indicator combination size for RMSPE when training and testing on ADC21.}
    \label{fig:BestSpearman21RMSPE}
\end{figure*}

\begin{figure*}
    \centering
    \includegraphics[width=0.75\linewidth]{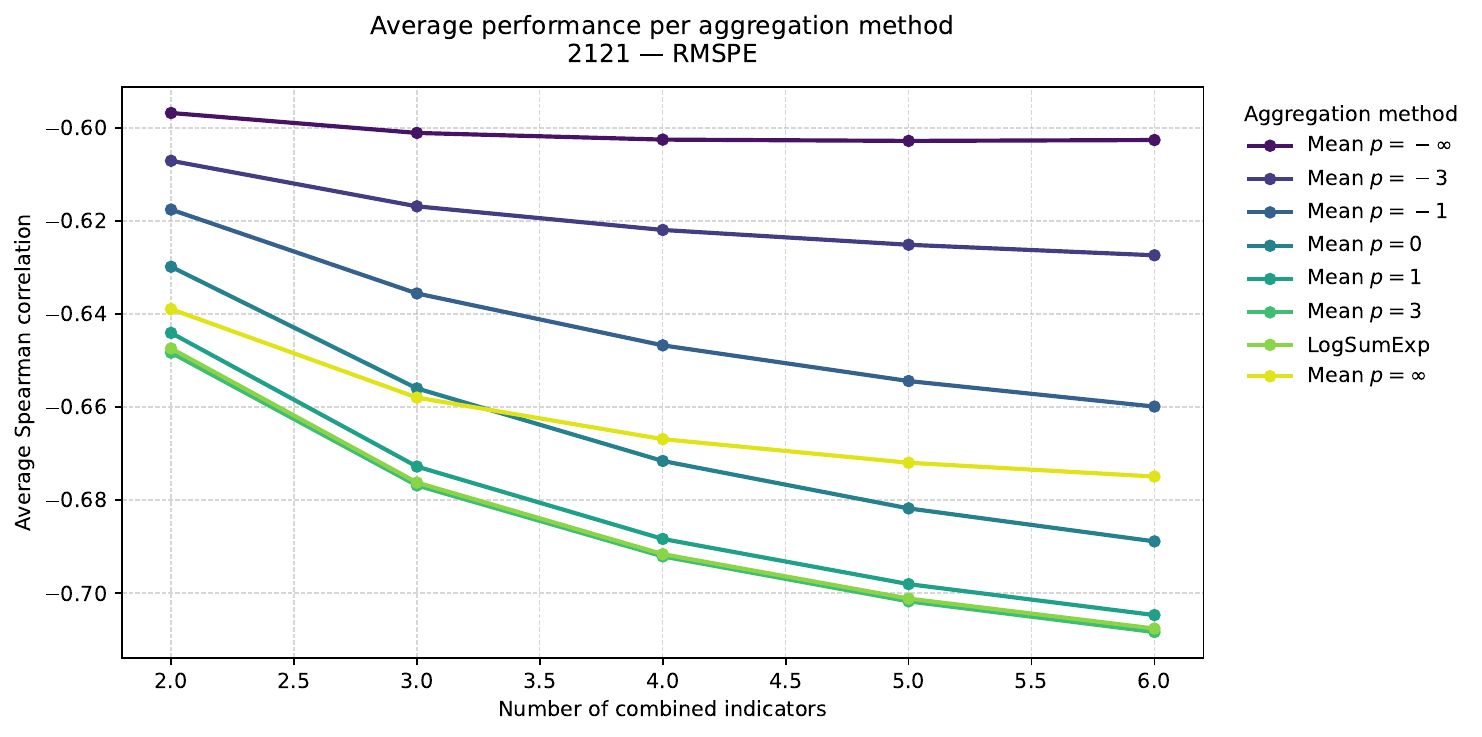}
    \caption{Average Spearman correlation per aggregation method as a function of indicator combination size for RMSPE when training and testing on ADC21.}
    \label{fig:AverageSpearman21RMSPE}
\end{figure*}

The performance of the safety cage framework can now be comprehensively evaluated using the operational coverage–error curves, as the best mathematical formulations for indicator fusion and the safety indicator per family have been established. For the combined RMSPE analysis illustrated in \autoref{fig:FinalRMSPE}, the most reliable operational curves are achieved through Indicator Fusion, Influence Functions, and Ensemble Consistency. These three approaches demonstrate superior behaviour across both internal and cross-domain datasets. Unlike alternative metrics, these three approaches demonstrate superior stability, avoiding the drastic error spikes that signify the accidental acceptance of high-error anomalies. This steady downward trajectory from right to left ensures that the samples identified as the absolute safest are indeed high-accuracy predictions across both in-domain and cross-domain datasets. For the combined SAM analysis presented in \autoref{fig:FinalSAM}, the best-performing strategies shift slightly to Uncertainty Quantification, Indicator Fusion, and Influence Functions. While UQ yields the lowest overall error profile on the in-domain ADC21 dataset, it exhibits a notable vulnerability at the start of the inclusion process, where the initial error curve decreases sharply. This trend of early misidentification persists during cross-domain validation on ADC19. By contrast, both Indicator Fusion and Influence Functions prove to be significantly more robust as they correctly prioritise the highest-fidelity samples without the initial performance degradation observed in the UQ setup. Consequently, while UQ remains a strong contender for stationary, in-domain data, the monotonic and predictable behaviour of the fusion and influence-based metrics provides a far more dependable foundation for defining operational safety boundaries under shifting environmental conditions. Overall, the evaluation establishes indicator fusion as the optimal deployment strategy, followed closely by influence functions, and then ensemble consistency.

\subsection{Combinations and Aggregation Analysis}\label{sub:ICAA}
While the preceding analysis explored fusion strategies across all six indicator families to exploit their complementary information, this section presents a systematic evaluation of all possible combinations between these families and their aggregation methods. This combinatorial approach enables the identification of configurations that maintain strong operational performance with a reduced computational cost, a critical consideration under strict runtime constraints.

For each subset of size two to six, aggregated risk scores were constructed using the operators introduced previously, namely the generalised means with varying p-values and LogSumExp. Each configuration was evaluated using the Spearman correlation between the aggregated risk score and the true error, quantifying the ability of the indicator to rank samples by risk. In total, 456 configurations were assessed. Utilising the Spearman correlation coefficient enables this combinatorial analysis to be executed with minimal computational overhead, requiring only a few seconds to evaluate the entire search space. By contrast, deriving the full Cov-Opt metric introduces a severe computational bottleneck. Because the evaluation must simulate a sequential deployment pipeline, the aggregated safety score must be re-computed every time a new sample is introduced to the operational domain. This process amounts to 8510 incremental evaluation loops for each of the 456 candidate configurations. This makes Spearman correlation a practical and efficient tool for systematically exploring the space of indicator combinations and aggregation methods. As before, a higher negative Spearman correlation indicates better performance of the risk indicator.

The results are first visualised through heatmaps (\autoref{fig:heatmap21RMSPE}, \autoref{fig:heatmap19RMSPE}, \autoref{fig:heatmap21SAM}, \autoref{fig:heatmap19SAM}), where each row corresponds to an indicator combination and each column to an aggregation method. Different patterns emerge from this analysis. For RMSPE, combinations involving Uncertainty Quantification, Influence Functions, and Ensemble Consistency systematically achieve stronger correlations, whereas combinations that rely more heavily on Out-of-Domain Detection, Activation Pattern Monitoring, or Adversarial Attack indicators exhibit weaker and less stable behaviour. For SAM, this pattern is seen for combinations involving Uncertainty Quantification and Influence Functions. This indicates that the indicators do not contribute equally, and that a subset of signals can provide more reliable information on model degradation. From the heatmaps, it is further observed that aggregation methods with higher p-values, in particular p=1, p=3, and LogSumExp, consistently outperform those with lower p-values. This indicates that emphasising the largest individual risk signal improves performance, supporting the interpretation that a single failing safety check should be a reason for concern.

To further analyse this effect, \autoref{fig:BestSpearman21RMSPE}, \autoref{fig:AverageSpearman21RMSPE}, \autoref{fig:BestSpearman19RMSPE}, \autoref{fig:AverageSpearman19RMSPE}, \autoref{fig:BestSpearman21SAM}, \autoref{fig:AverageSpearman21SAM}, \autoref{fig:BestSpearman19SAM}, and \autoref{fig:AverageSpearman19SAM} present the average and best achievable performance per aggregation method as a function of the combination size, where a less negative correlation value corresponds to a deterioration in overall performance. For RMSPE, increasing the combination size generally degrades the best achievable performance for negative p-values, such as p=-$\infty$. Conversely, this effect is less pronounced for p=1, p=3, and LogSumExp. These aggregation methods consistently maintain stronger performance across combination sizes, reinforcing the trends observed in the heatmaps. In contrast, the average performance improves with increasing combination size. This can be attributed to the presence of poorly performing small combinations, which negatively impact the average, while larger combinations benefit from incorporating multiple complementary sources of risk, leading to more robust overall behaviour. For SAM, similar trends are observed for both the best performance and the average performance. However, in this case, the best achievable performance decreases with increasing combination size across all aggregation methods. This suggests that, for this metric, adding additional indicators introduces redundancy or noise that limits the maximum attainable ranking quality.

\begin{figure*}
    \centering
    \includegraphics[width=0.8\linewidth]{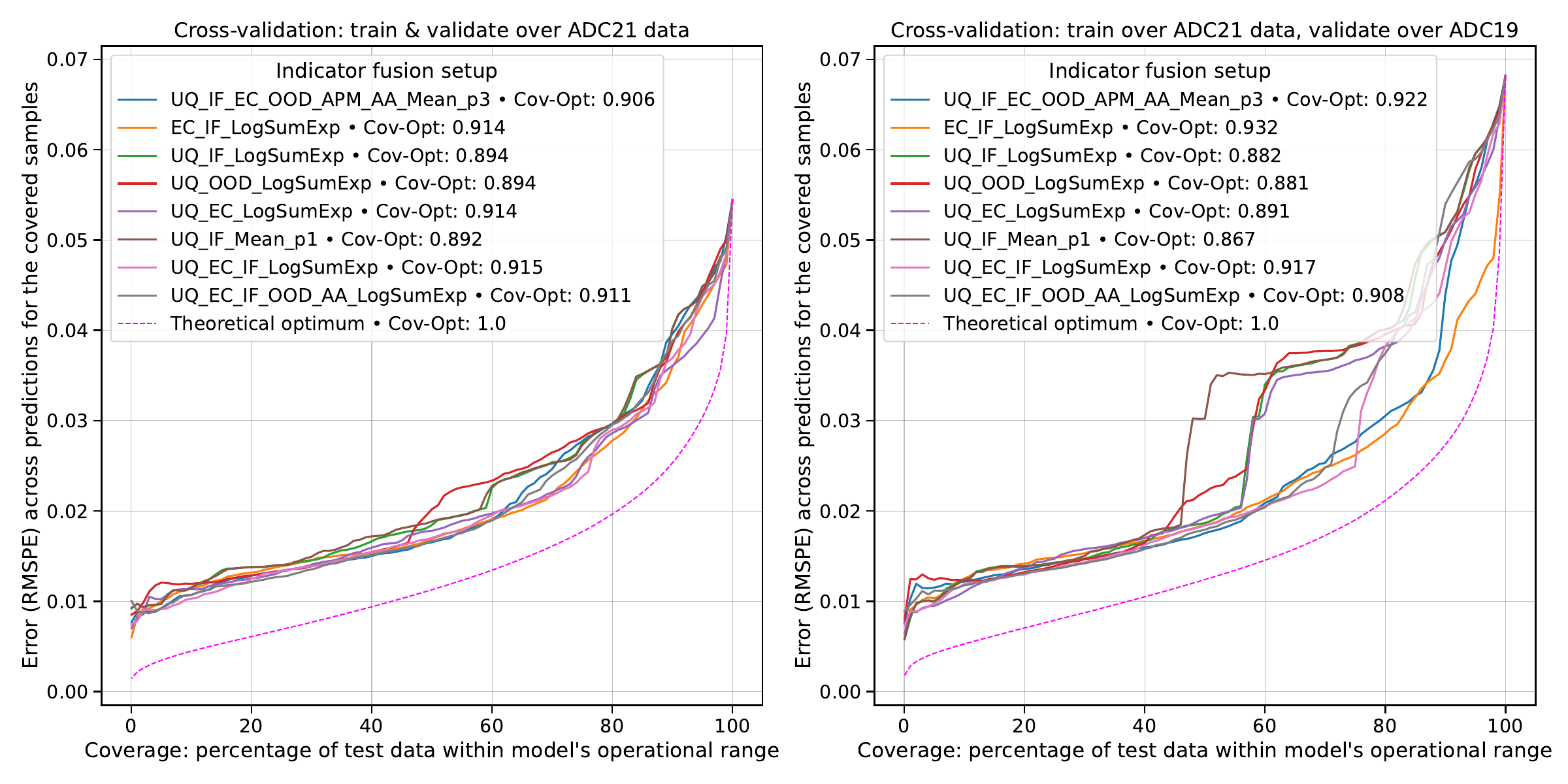}
    \caption{Illustration of RMSPE evolution as samples are admitted to the model's operational range, based on different possible best combinations according to the previous analyses, comparing internal (ADC21) and cross-domain (ADC19) validation.}
    \label{fig:CombAggrRMSPE}
\end{figure*}

\begin{figure*}
    \centering
    \includegraphics[width=0.8\linewidth]{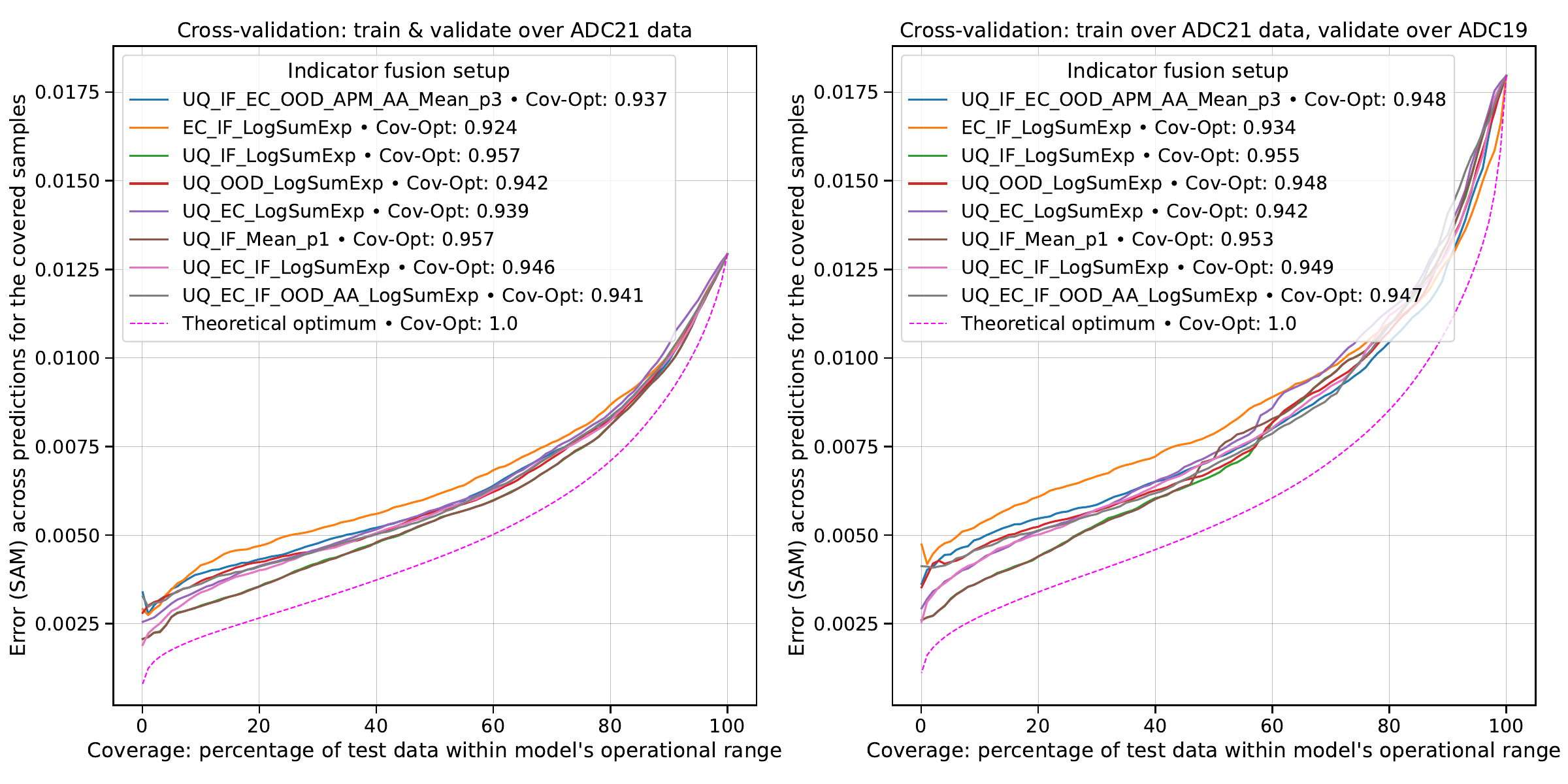}
    \caption{Illustration of SAM evolution as samples are admitted to the model's operational range, based on different possible best combinations according to the previous analyses, comparing internal (ADC21) and cross-domain (ADC19) validation.}
    \label{fig:CombAggrSAM}
\end{figure*}

However, correlation alone is not the final objective. The ultimate goal is to obtain a safety indicator that yields optimal behaviour in the coverage–error trade-off, quantified through the Cov-Opt metric. The coverage–error plots presented in \autoref{fig:CombAggrRMSPE} and \autoref{fig:CombAggrSAM} provide this evaluation. In this plot, the best fusion strategy including all six indicators is shown (UQ\_IF\_EC\_OOD\_APM\_AA\_Mean\_p3) as well as the combinations with the two best indicators per case (EC\_IF\_LogSumExp for RMSPE and UQ\_IF\_LogSumExp for SAM). In addition, a combination with the most discussed methods in literature is added (UQ\_OOD\_LogSumExp), and one that combines two indicators that quantify uncertainty in different ways (UQ\_EC\_LogSumExp). Finally the best combinations from the previous analysis are added (UQ\_IF\_Mean\_p1 for SAM, UQ\_EC\_IF\_LogSumExp for RMSPE 2121, and UQ\_EC\_IF\_OOD\_AA\_LogSumExp for RMSPE 2119).

These results indicate that configurations that achieve strong correlation generally translate into strong coverage–error performance. For the ADC21 setting, the combinations identified as optimal based on correlation also correspond to those achieving the highest Cov-Opt values for both RMSPE and SAM. For ADC19, the correspondence is less direct, although the configurations selected through correlation analysis still achieve high Cov-Opt values, consistently exceeding 0.9. In the case of RMSPE, the highest Cov-Opt is obtained when combining EC and IF, and the second highest when combining all available indicators, rather than the subset of five suggested by the correlation analysis. This shows that combining the two best indicators for this scenario gives strong Cov-Opt, while combining all indicators can further improve performance compared to when only using the subset of five from the correlation analysis. For SAM, the optimal combination of indicators remains consistent with the correlation-based selection, although the best performance is achieved using a different aggregation method, further highlighting the sensitivity of the final performance to the choice of aggregation. It should, however, be noted that these performance variations across the different aggregation methods remain marginal. Overall, these results confirm that correlation-based analysis provides a strong and reliable proxy for identifying effective configurations and understanding the coverage pattern, while also emphasising that the final performance depends on both the indicator set and the aggregation strategy. Combinations of the two best indicators per case consistently give strong performance across all four cases, and are the best option in three of them. For OOD, the configurations including this indicator do not yield the best results in any case, indicating that out-of-domain detection is insufficient to identify all instances of degraded model performance.

These findings directly inform the final safety cage design, where a limited set of high-quality indicators is combined using a max-like aggregation strategy on risk to achieve good coverage–error performance. This shows that it is possible to maximise the safety cage performance while taking into account the complexity of the model and the computational costs. In addition, this analysis serves to characterise the behaviour of indicator combinations and aggregation strategies rather than to define a universally optimal configuration. The results are inherently dependent on the problem setting, including the underlying model, data distribution, and evaluation metric, as shown by the observed differences between SAM and RMSPE. As such, the selection of indicator combinations and aggregation methods should be treated as a model- and task-specific procedure, and repeated when transitioning to new datasets, models, or operational conditions, as well as before deployment.

\section{Conclusions}
This work investigated the development of a safety cage framework capable of identifying unreliable predictions in ML models used to obtain exoplanet atmospheric transmission spectra. While modern ML models can achieve high predictive performance, their deployment in operational pipelines requires mechanisms to detect situations in which predictions may be unreliable. Ensuring that such models can provide reliable outputs, or signal when they cannot, is therefore essential for their safe integration into scientific pipelines.

To address this challenge, six complementary classes of safety indicators were analysed, including anomaly detection techniques, uncertainty quantification methods, ensemble disagreement metrics, adversarial robustness measures, influence-based indicators, and activation-pattern monitoring. Each of these indicators captures a different aspect of model behaviour and potential failure modes. Their individual performance was evaluated using correlation analysis with prediction error metrics and through operational metrics from the coverage-error curves. The results show that individual indicators provide meaningful signals regarding prediction reliability, but their effectiveness varies depending on the error metric and dataset. No single indicator consistently dominates across all scenarios, confirming that model failures can arise from multiple sources. This observation motivates the use of indicator fusion, where multiple safety signals are aggregated to produce a single reliability score.

Several fusion strategies were investigated, including LogSumExp and a family of generalised means with different power parameters. The analysis demonstrates that aggregation methods that place greater emphasis on higher risk indicators produce stronger correlations with prediction errors and more favourable coverage-error trade-offs. In particular, the generalised mean with $p = 3$ and the LogSumExp aggregation consistently achieve the best performance across both datasets and error metrics. Based on this analysis, the generalised mean with $p = 3$ was selected as the primary fusion strategy due to its balance between sensitivity to elevated risk signals and stability across datasets.

Operationally, the fused safety score allows the system to filter unreliable predictions with minimal loss of coverage. For example, when using the Mean\_p3 fusion strategy, a reduction of approximately 20\% in coverage results in error reductions of around 45\% in RMSPE for ADC21 and more than 65\% for ADC19. Similar improvements are observed for the SAM metric, with reductions of approximately 35\% for ADC21 and 40\% for ADC19. In practical terms, a relatively small loss in coverage leads to a substantial improvement in prediction reliability, highlighting the effectiveness of the proposed indicator fusion strategy for operational deployment.

In general, this study demonstrates that combining multiple monitoring signals into a unified safety indicator provides a robust mechanism to detect unreliable model predictions. By integrating complementary metrics that capture distinct failure modes, the proposed safety cage framework overcomes the limitations of single-indicator monitoring, allowing machine learning architectures to operate with increased transparency and reliability. Crucially, the introduction of the Cov-Opt metric provides a scalable, systematic methodology for identifying the optimal indicator configurations without prohibitive computational overhead. Because this framework remains structurally invariant across diverse data regimes and modelling approaches, it represents an important step toward the reliable, safe deployment of data-driven models in high-stakes scientific domains, such as exoplanet atmospheric retrieval.

\section*{Acknowledgements}
Research funded by the ESA project ``Machine Learning and Artificial Intelligence Algorithms for Exoplanet Atmospheres Detection and Analysis'', Contract No. 4000148493/25/NL/KG.

\section*{Data Availability}
The datasets used in this study are publicly available on Zenodo \cite{yip_2025_15050868}. The data were produced as part of the Ariel Data Challenge. The repository provides documentation on the datasets, variables, and guidelines for using the data.

\bibliographystyle{rasti}
\bibliography{references}



\appendix

\section{Selection of Safety Indicators}\label{App:safetyInd}
This appendix provides a comprehensive technical record of the evaluation, derivation, and final selection process for the safety indicators.

\subsection{Anomaly Detection}
\begin{figure*}
    \centering
    \includegraphics[width=0.8\linewidth]{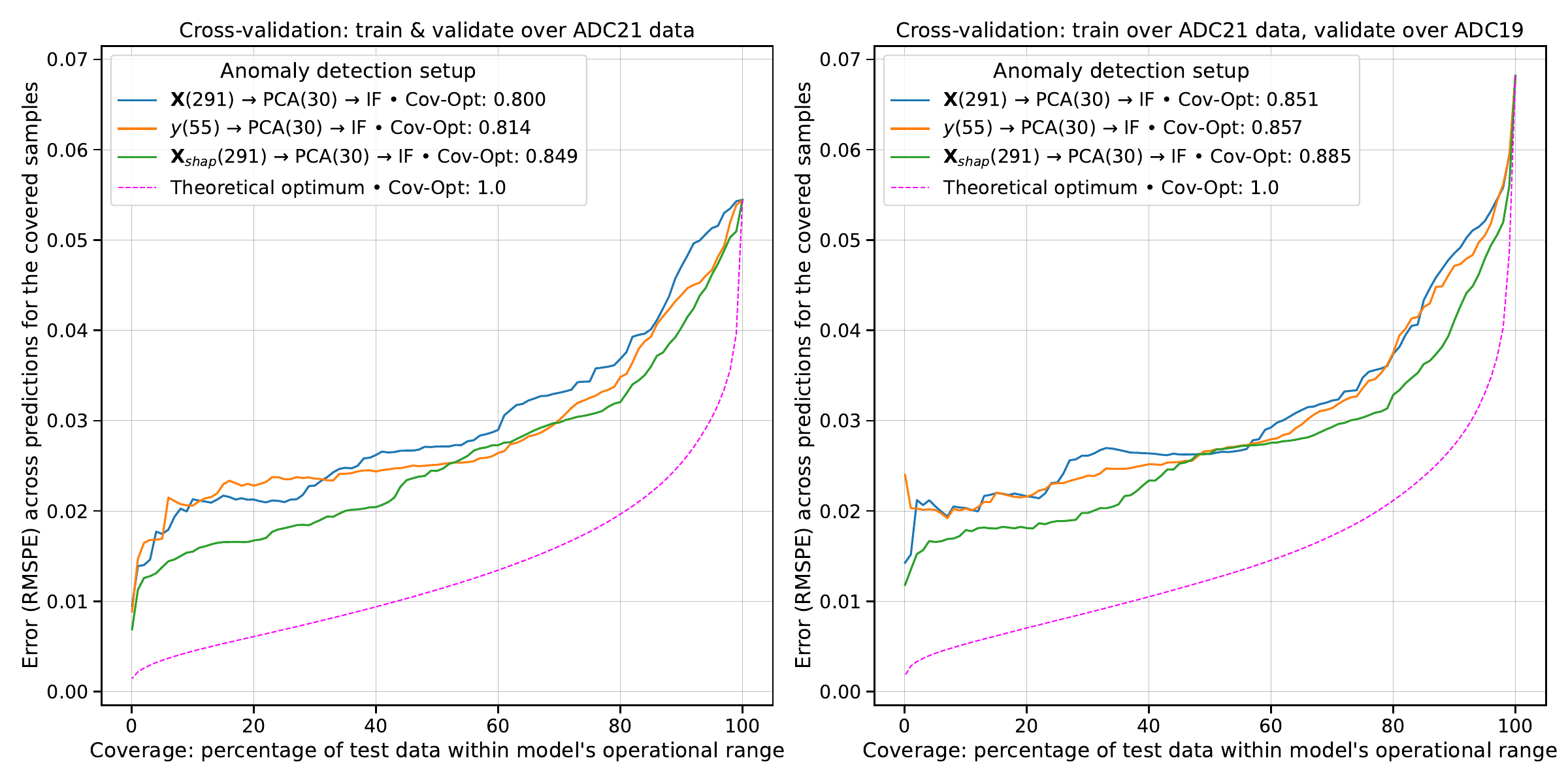}
    \caption{Illustration of RMSPE evolution as samples are admitted to the model's operational range, based on their safety scores for \textbf{Anomaly Detection} using X, y, and $\text{X}_{\text{shap}}$. The x-axis (Coverage) represents the iterative process of starting with only the safest data points and progressively lowering the safety threshold until 100\% of the test data has been included. The left panel displays results for a model trained and tested on ADC21 data, while the right panel shows results for a model trained on ADC21 and tested on the ADC19 dataset.}
    \label{fig:OODRMSPE}
\end{figure*}

\begin{figure*}
    \centering
    \includegraphics[width=0.8\linewidth]{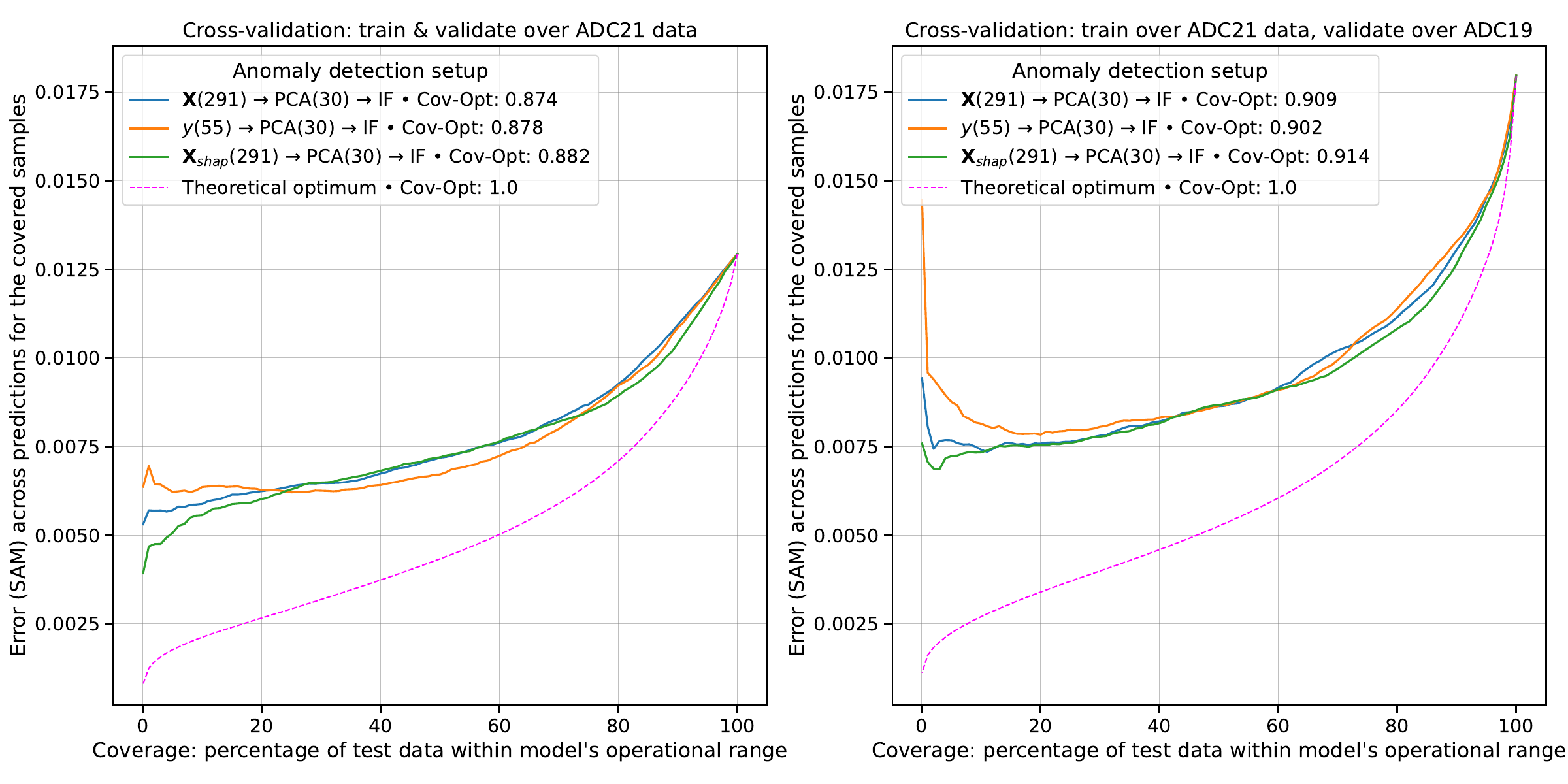}
    \caption{Illustration of SAM evolution as samples are admitted to the model's operational range, based on their safety scores for \textbf{Anomaly Detection} using X, y, and $\text{X}_{\text{shap}}$, following the approach in \autoref{fig:OODRMSPE}. The x-axis represents the progression from the safest data points to 100\% test data inclusion.}
    \label{fig:OODSAM}
\end{figure*}

To evaluate the effectiveness of the proposed OOD Detection framework, three distinct safety indicators are analysed: X, y, and $\text{X}_{\text{shap}}$. These indicators rank the reliability of model predictions, where a higher safety score indicates greater confidence that the input remains within the model's learned operational range.

In the OOD Detection setup, anomaly detection is performed using Isolation Forest models to determine if a sample resides in a low-density region of the training data. Samples located in such regions indicate that the predictive model has insufficient information to provide a reliable result. Due to the high dimensionality of the dataset with 291 input features and 55 targets, the data is projected into a 30-dimensional space using PCA, an approach following the methodology presented in \cite{safetycages}. The specific data projected and monitored depends on the chosen safety indicator. The \textbf{$\mathbf{X}$ }indicator performs anomaly detection directly on the input feature space. This metric assesses whether the current input vector aligns with the distribution of the training data, thereby identifying samples that are geometrically distant from the established operational range. The \textbf{$\mathbf{y}$} indicator monitors the model's output space by applying anomaly detection to the predicted values. This approach identifies predictions that fall outside the expected statistical range seen during the training phase. The \textbf{$\text{X}_{\text{shap}}$} indicator operates within the explanation space, utilising SHAP values derived from a linear explainer using a Ridge model \citep{lundberg2017unified}. Note, the predictive model remains ELM. By encoding the input as the median SHAP values across spectral dimensions, this indicator monitors whether the underlying feature explanation patterns of a prediction remain consistent with the logic learned from the training data.

The relationship between these indicators and prediction performance is assessed through Spearman correlation coefficients and coverage-error curves. As shown in \autoref{tab:OODCorr}, $\text{X}_{\text{shap}}$ demonstrates the best overall behaviour and the strongest negative correlation with RMSPE, and y with SAM. It can also be seen that, in general, there is an increase in correlation when moving to cross-domain testing, showing that this metric does not degrade with data drift.

\autoref{fig:OODRMSPE} and \autoref{fig:OODSAM} illustrate that by admitting samples starting with the highest safety scores, significantly lower initial errors are achieved. The $\text{X}_{\text{shap}}$ indicator, represented by the green curve, maintains, in general, the lowest error profile throughout the coverage range in both internal validation and cross-domain validation, and has, for all scenarios, the highest Cov-Opt score.

Given its superior ability to identify high-error samples before they are admitted into the operational range, $\text{X}_{\text{shap}}$ is identified as the most robust safety indicator for Anomaly Detection. For both the RMSPE and SAM curves, $\text{X}_{\text{shap}}$ consistently remains below its counterparts, showing a controlled increase in error as the safety threshold is relaxed to increase coverage. The indicator retains its predictive power even when moving from ADC21 to ADC19 data, demonstrated by a higher correlation and Cov-Opt score in the presence of data drift.

\begin{table}
    \centering
    \caption{Spearman correlation coefficients between the proposed safety indicators (X, y, and $\text{X}_{\text{shap}}$) for Anomaly Detection and the prediction error metrics (RMSPE and SAM). Results are shown for internal validation (ADC21) and cross-domain validation (ADC19). Negative values indicate that higher safety scores effectively correspond to lower prediction errors.}
    \begin{tabular}{c|c|c|c|c}
    \hline
      Indicator & RMSPE 21 & RMSPE 19 & SAM 21 & SAM 19\\ \hline
     $X$  & -0.512 & -0.537 & -0.489 & -0.492\\
     $Y$  & -0.488 & -0.484 &  -0.515 & -0.518\\
     $X_{shap}$ & -0.548 & -0.576 &  -0.498 & -0.504 \\ \hline
    \end{tabular}
    \label{tab:OODCorr}
\end{table}

Comparing these results with those presented in \cite{safetycages}, which also investigates anomaly detection to bound the operational range using Ridge regression as the underlying model, reveals consistent trends. In both studies, the same monitoring techniques were applied to the same indicator variables (X, y, and $\text{X}_{\text{shap}}$). The results show that $\text{X}_{\text{shap}}$ remains the best-performing indicator even when moving to a more complex modelling approach based on ELM. While the relative performance differences between the indicators are smaller in the present study, this can be explained by the improved predictive accuracy of the ELM model, which results in more reliable $y_{\text{pred}}$ values.

\subsection{Uncertainty Quantification}
\begin{figure*}
    \centering
    \includegraphics[width=0.8\linewidth]{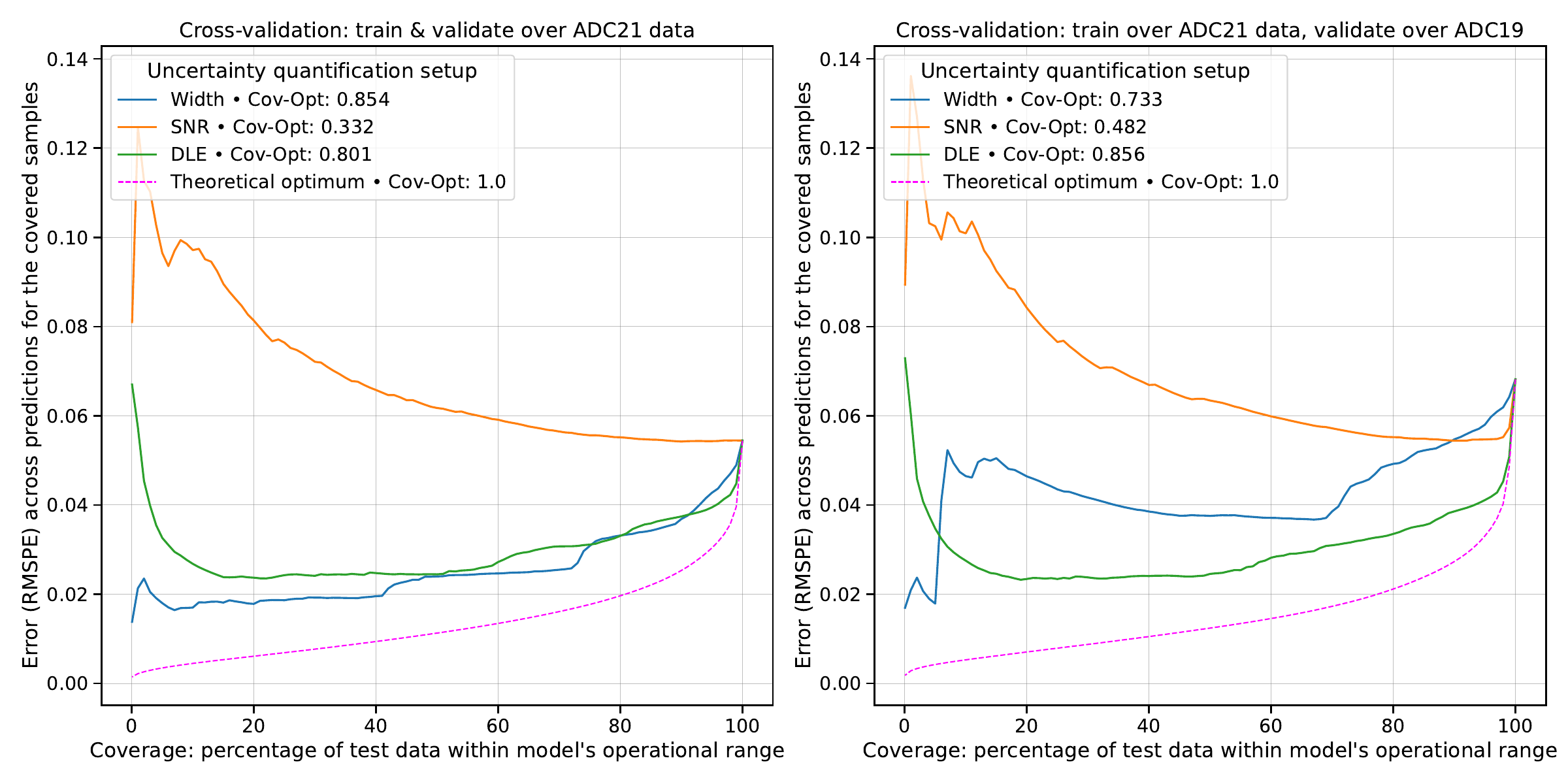}
    \caption{Illustration of RMSPE evolution as samples are admitted to the model's operational range, based on their safety scores for \textbf{Uncertainty Quantification} using Width, SNR, and DLE, following the approach in \autoref{fig:OODRMSPE}. The x-axis represents the progression from the safest data points to 100\% test data inclusion.}
    \label{fig:UQRMSPE}
\end{figure*}

\begin{figure*}
    \centering
    \includegraphics[width=0.8\linewidth]{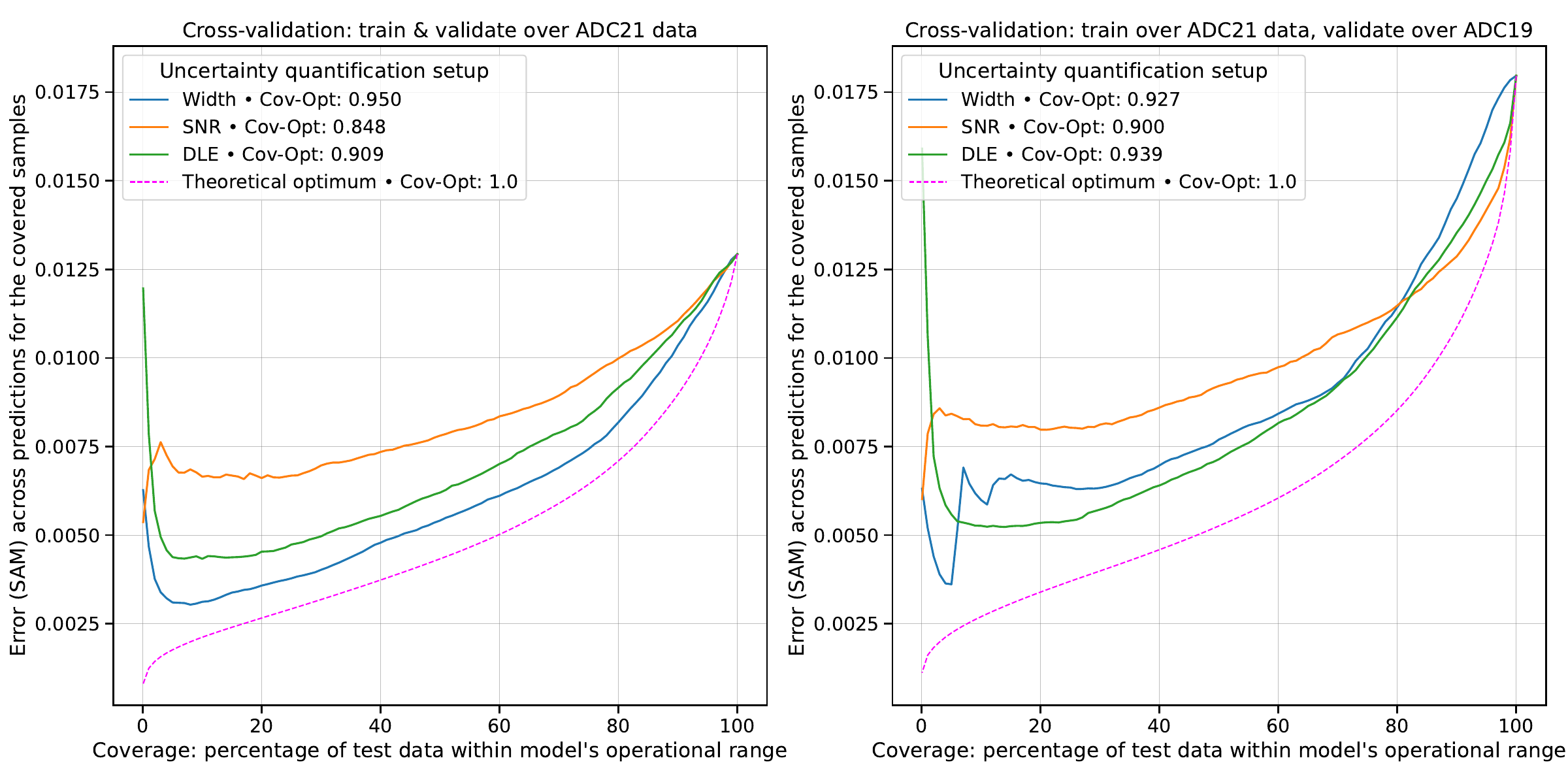}
    \caption{Illustration of SAM evolution as samples are admitted to the model's operational range, based on their safety scores for \textbf{Uncertainty Quantification} using Width, SNR, and DLE, following the approach in \autoref{fig:OODRMSPE}. The x-axis represents the progression from the safest data points to 100\% test data inclusion.}
    \label{fig:UQSAM}
\end{figure*}

In the UQ setup, three indicators, Width, SNR, and DLE, are evaluated for their ability to rank the uncertainty of predictions. Following the same analytical framework established for OOD Detection, the effectiveness of these indicators is assessed by the correlation with error metrics and the performance across expanding coverage levels.

The Direct Loss Estimation (\textbf{DLE}) indicator is based on a wrapper that explicitly models the residuals of the base system. The safety model is trained to predict the magnitude of error ($r_s$) based on the input features and the initial prediction. In addition to utilising an ELM to estimate expected residuals, conformal prediction is applied to obtain an interval with guaranteed coverage. To establish the safety boundaries for DLE, the predicted residual and the conformity score for the residuals are added to and subtracted from the predicted output to obtain the lower and upper bounds. During the inference phase, the safety score is calculated as the negative mean of the predicted interval width relative to the prediction itself across 55 wavelengths. This approach identifies samples where the expected error is disproportionately high compared to the signal, marking them as less safe. The negative sign is a convention to ensure that a higher safety score consistently represents a safer prediction and is also applied to multiple indicators described below. The \textbf{Width} indicator utilises a more complex architecture combining PCA and Quantile Regression (QR). By projecting error terms into a lower-dimensional PCA space, the system trains a Multi-Output Regressor using LightGBM \citep{ke2017lightgbm} to estimate the 5th and 95th percentiles of the principal components obtained from PCA. This use of PCA reduces the dimensionality of the 55-channel output vector and accounts for the correlation between wavelength channels, as individual molecules typically influence multiple channels simultaneously. Next, the lower and upper prediction bounds are established. These are calculated by combining the quantiles with conformity scores from a calibration set, which is generated via nested cross-validation (where an inner cross-validation loop runs on each training fold). This conformity score is calculated based on the maximum error between either the lower bound and true calibration value or the true calibration value and the upper bound, effectively accounting for the worst-case deviation. The final safety score is defined as the negative mean of the resulting conformal interval width ($y_{high} - y_{low}$) divided by the prediction itself across all wavelengths. Consequently, a wider relative interval indicates greater model uncertainty and lower prediction reliability. The Signal-to-Noise Ratio (\textbf{SNR}) indicator provides a relative measure of certainty by comparing the peak-to-peak amplitude of the spectral prediction, which is a function of the planet's atmospheric scale height, to its estimated noise. Scale height is the characteristic vertical thickness of the planet's atmospheric boundary, dictating the total physical size of the spectral signal and directly determining how visible and clear those atmospheric features will appear to a telescope. In this context, the noise is represented by half of the mean width obtained through the QR method above. By calculating the ratio of the predicted signal range to this uncertainty measure, the system distinguishes between high-confidence predictions (high SNR) and those where the estimated error is large enough to obscure the underlying astrophysical signal (low SNR).

\begin{table}
    \centering
    \caption{Spearman correlation coefficients between the proposed safety indicators for Uncertainty Quantification and the prediction error metrics, following the experimental setup in \autoref{tab:OODCorr}.}
    \begin{tabular}{c|c|c|c|c}
    \hline
      Indicator   & RMSPE 21 & RMSPE 19 & SAM 21 & SAM 19\\ \hline
     Width  &  -0.667 & -0.674 & -0.799 & -0.777\\
     SNR & -0.228 & -0.278 & -0.542 & -0.580 \\
     DLE  & -0.582 & -0.622 & -0.687 & -0.707\\ \hline
    \end{tabular}
    \label{tab:UQCorr}
\end{table}

As shown in \autoref{tab:UQCorr}, the Width indicator exhibits the strongest and most consistent negative correlation with both RMSPE and SAM across all datasets. It achieves particularly high values for the SAM metric, reaching $-0.799$ on ADC21 and $-0.777$ on ADC19. The DLE indicator also demonstrates strong performance, with correlations ranging from $-0.582$ to $-0.707$. In contrast, the SNR indicator shows the weakest relationship with prediction error, especially for the RMSPE metric on ADC21, where the correlation is only $-0.228$. For Width, there is a slight decrease in performance when moving from ADC21 to ADC19. However, this is just a slight decrease, and Width is still a good indicator in a cross-domain scenario.

The analysis of the coverage-error behaviour shown in \autoref{fig:UQRMSPE} and \autoref{fig:UQSAM} reveals distinct trends between internal and cross-domain validation. During internal validation on the ADC21 dataset, Width maintains a superior and more stable performance profile, achieving a higher Cov-Opt score. However, in the cross-domain validation on ADC19, DLE shows competitive performance as coverage increases. Despite this, a critical observation from the curves is that DLE tends to misidentify the initial safest cases, resulting in higher initial error spikes compared to Width, which better identifies the lowest-error samples from the onset.

Based on its superior correlation profile and its ability to consistently prioritise low-error samples without the initial misidentifications observed in DLE, Width is identified as the most robust indicator for this setup. It provides the most reliable trade-off between maximising data coverage and minimising prediction error, ensuring a smooth degradation of model performance as less certain samples are admitted to the operational range.

\subsection{Ensemble Consistency}
\begin{figure*}
    \centering
    \includegraphics[width=0.8\linewidth]{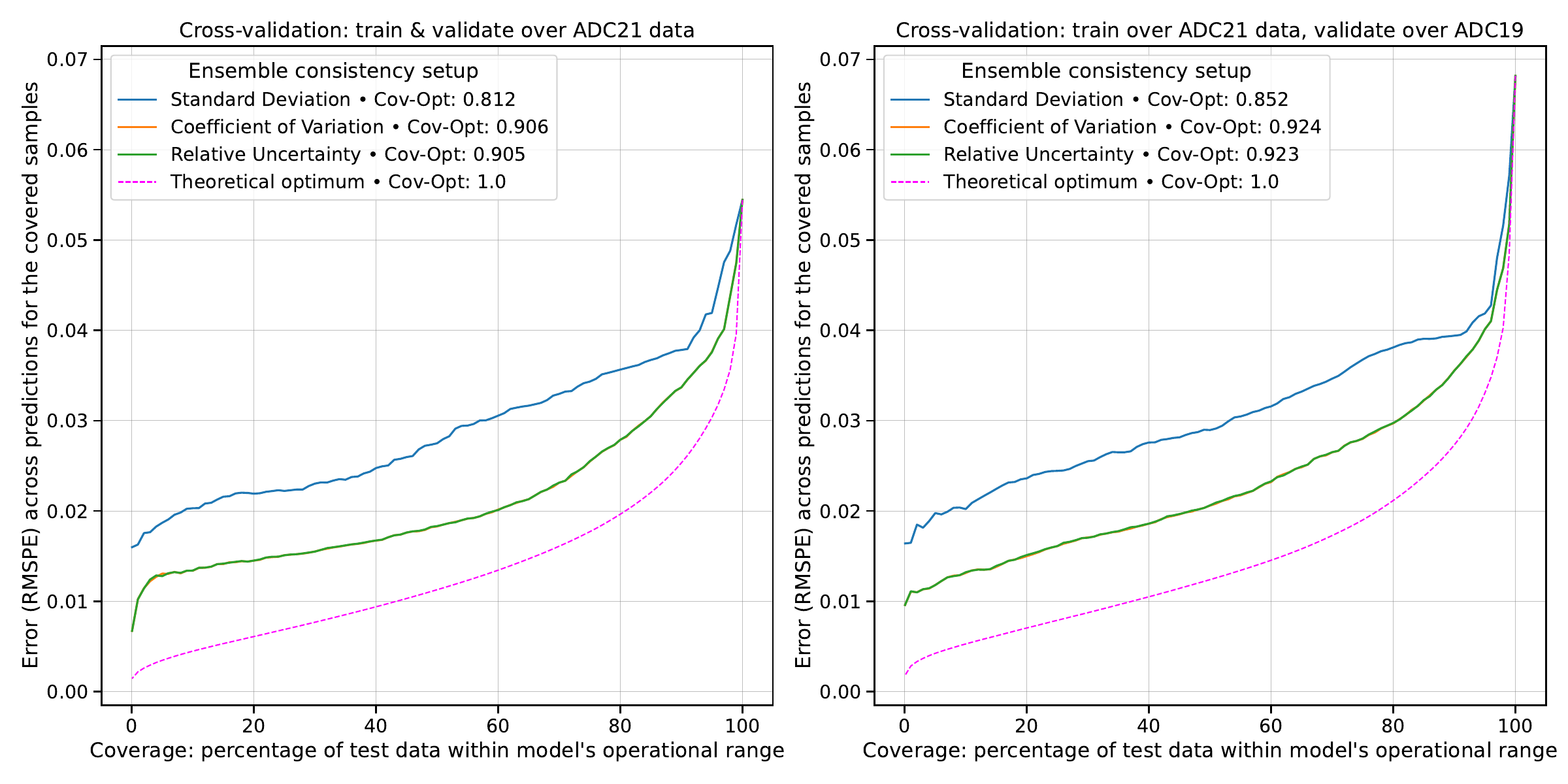}
    \caption{Illustration of RMSPE evolution as samples are admitted to the model's operational range, based on their safety scores for \textbf{Ensemble Consistency} using Standard Deviation, Coefficient of Variation, and Relative Uncertainty, following the approach in \autoref{fig:OODRMSPE}. The x-axis represents the progression from the safest data points to 100\% test data inclusion.}
    \label{fig:ECRMSPE}
\end{figure*}

\begin{figure*}
    \centering
    \includegraphics[width=0.8\linewidth]{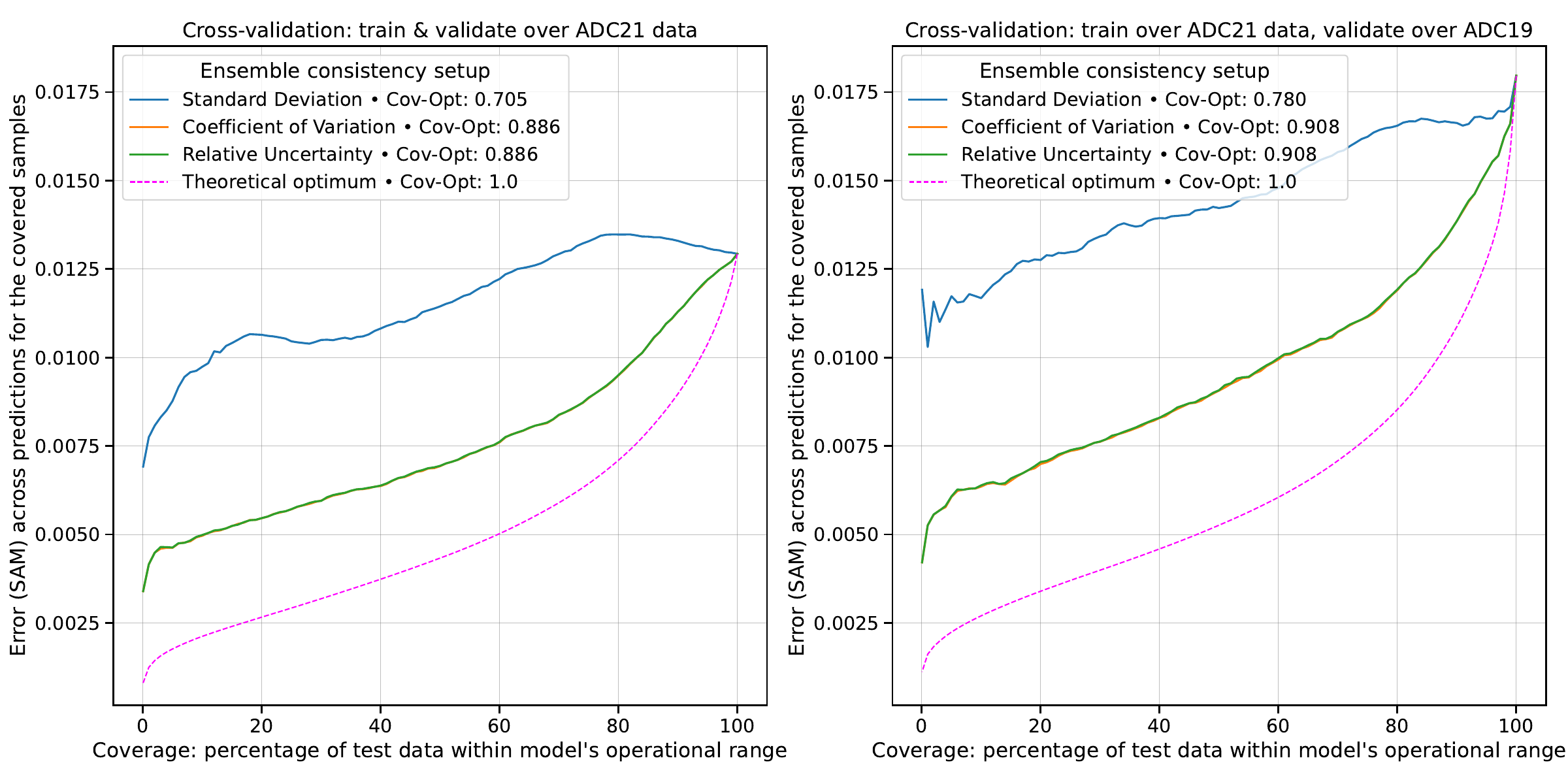}
    \caption{Illustration of SAM evolution as samples are admitted to the model's operational range, based on their safety scores for \textbf{Ensemble Consistency }using Standard Deviation, Coefficient of Variation, and Relative Uncertainty, following the approach in \autoref{fig:OODRMSPE}. The x-axis represents the progression from the safest data points to 100\% test data inclusion.}
    \label{fig:ECSAM}
\end{figure*}

In the Ensemble Consistency setup, three indicators, Standard Deviation, Coefficient of Variation, and Relative Uncertainty, are evaluated to rank the reliability of predictions. The assessment follows the same analytical framework as the previous sections, focusing on the correlation between indicators and error metrics across expanding coverage levels.

The \textbf{Standard Deviation} indicator quantifies the raw disagreement across the ensemble. For each input, predictions are generated by five models in the ensemble, and the standard deviation is calculated across these outputs for each wavelength. The difference between the five models is the data seen during training. The safety score is defined as the negative mean of this standard deviation across the 55 wavelengths. A larger spread in predictions results in a lower safety score, flagging the sample as potentially unreliable due to model inconsistency. The \textbf{Coefficient of Variation} indicator provides a relative measure of disagreement. It scales the ensemble standard deviation by the ensemble mean prediction. This normalisation ensures that the uncertainty is evaluated in the context of the signal magnitude, preventing high-amplitude signals from being unfairly penalised by naturally higher absolute variance. The safety score is the negative mean of this ratio across all spectral channels. The \textbf{Relative Uncertainty} further refines the coefficient of variation by incorporating information from the training phase. The coefficient of variation calculated during inference is divided by the average standard deviation observed during training for each wavelength. This benchmarks the current model inconsistency against the baseline disagreement inherent to the training data. If the current ensemble disagreement exceeds the typical levels seen during training, the safety score decreases, indicating that the input represents a scenario the ensemble is struggling to map consistently.

\begin{table}
    \centering
    \caption{Spearman correlation coefficients between the proposed safety indicators for Ensemble Consistency and the prediction error metrics, following the experimental setup in \autoref{tab:OODCorr}.}
    \begin{tabular}{c|c|c|c|c}
    \hline
      Indicator   & RMSPE 21 & RMSPE 19 & SAM 21 & SAM 19\\ \hline
     Standard Deviation  & -0.256 & -0.228 & 0.060 & 0.075\\
     Coefficient of Variation  & -0.632 & -0.600 & -0.531 & -0.480\\
     Relative Uncertainty  & -0.629 & -0.600 & -0.533 & -0.482 \\ \hline
    \end{tabular}
    \label{tab:ECCorr}
\end{table}

The Spearman correlation coefficients in \autoref{tab:ECCorr} reveal that Coefficient of Variation and Relative Uncertainty are the most effective indicators in this setup. Both demonstrate consistent negative correlations of around $-0.6$ for RMSPE and $-0.5$ for SAM across both ADC21 and ADC19 datasets. In contrast, Standard Deviation proves to be a highly unreliable indicator for this problem. It shows a weak negative correlation for RMSPE and, critically, a positive correlation with the SAM metric on both datasets. This positive correlation suggests that, for SAM, a decrease in safety does not reliably correspond to an increase in prediction error, making it unsuitable for defining safety boundaries.

The coverage-error curves in \autoref{fig:ECRMSPE} and \autoref{fig:ECSAM} further justify the selection of Relative Uncertainty or Coefficient of Variation over Standard Deviation. For both RMSPE and SAM, the Relative Uncertainty and Coefficient of Variation indicators overlap almost perfectly, maintaining the lowest error profile across the entire coverage range. The Standard Deviation indicator shows a significantly higher error from the onset and, in the cross-domain validation for SAM on ADC19, displays a sharp initial error spike. This indicator maintains a much higher error plateau than the other two metrics until roughly 95\% coverage is reached.

Both Coefficient of Variation and Relative Uncertainty demonstrate remarkable stability by identifying the lowest-error samples first and showing a gradual, controlled increase in error as coverage approaches 100\%, even in cross-domain scenarios. Because Relative Uncertainty and Coefficient of Variation avoid the misidentifications and high error plateaus associated with Standard Deviation, they are identified as the most robust metrics for this setup. Given the nearly identical performance of these two indicators, either serves as a reliable primary metric for defining operational safety boundaries based on ensemble consistency. However, due to the added standard deviation of the ensemble models during training, the Relative Uncertainty is chosen as the safety indicator to account for a possible case where the models themselves disagree more with each other.

\subsection{Adversarial Attack}
\begin{figure*}
    \centering
    \includegraphics[width=0.8\linewidth]{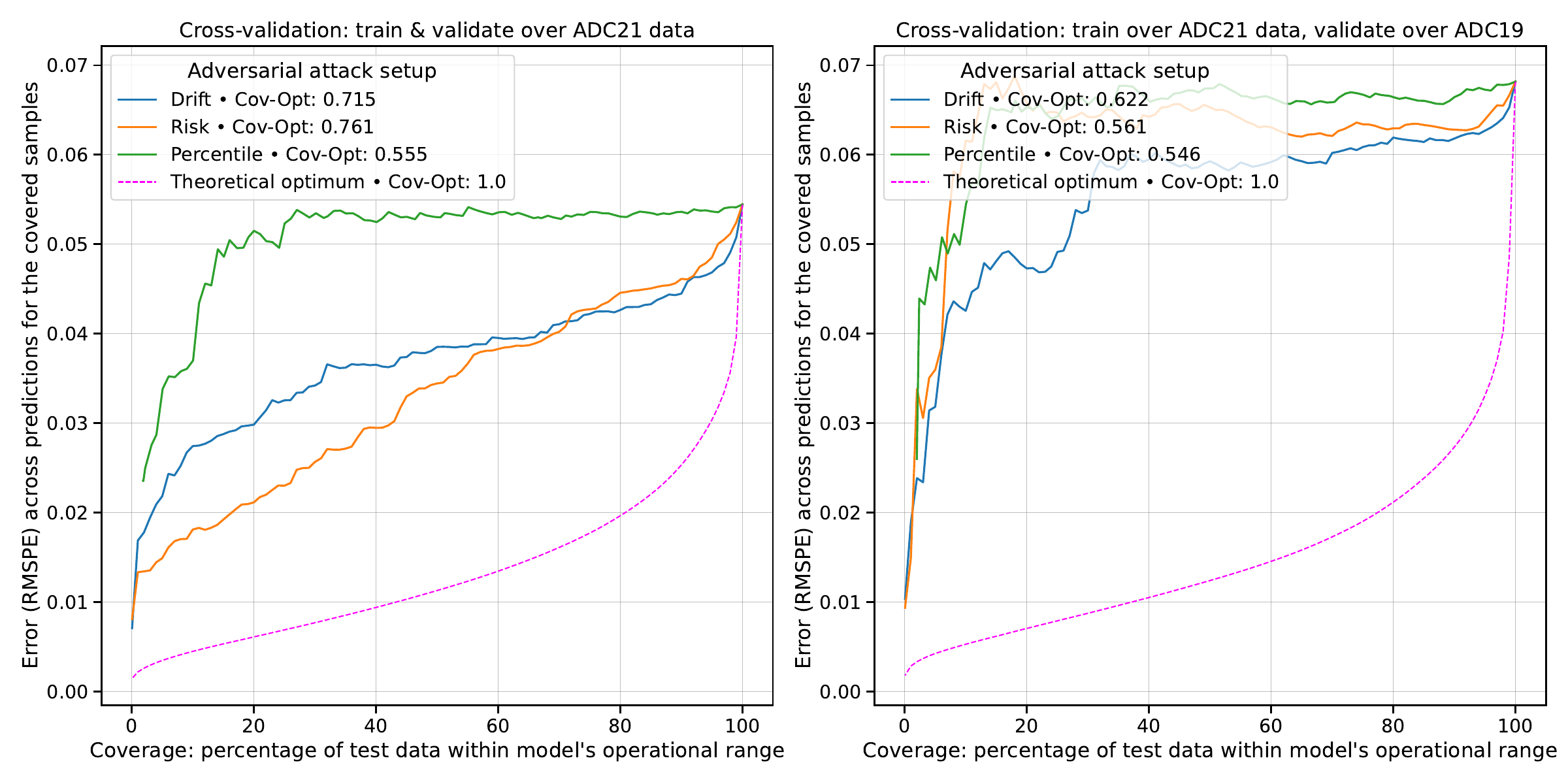}
    \caption{Illustration of RMSPE evolution as samples are admitted to the model's operational range, based on their safety scores for \textbf{Adversarial Attack} using Drift, Risk, and Percentile, following the approach in \autoref{fig:OODRMSPE}. The x-axis represents the progression from the safest data points to 100\% test data inclusion.}
    \label{fig:AARMSPE}
\end{figure*}

\begin{figure*}
    \centering
    \includegraphics[width=0.8\linewidth]{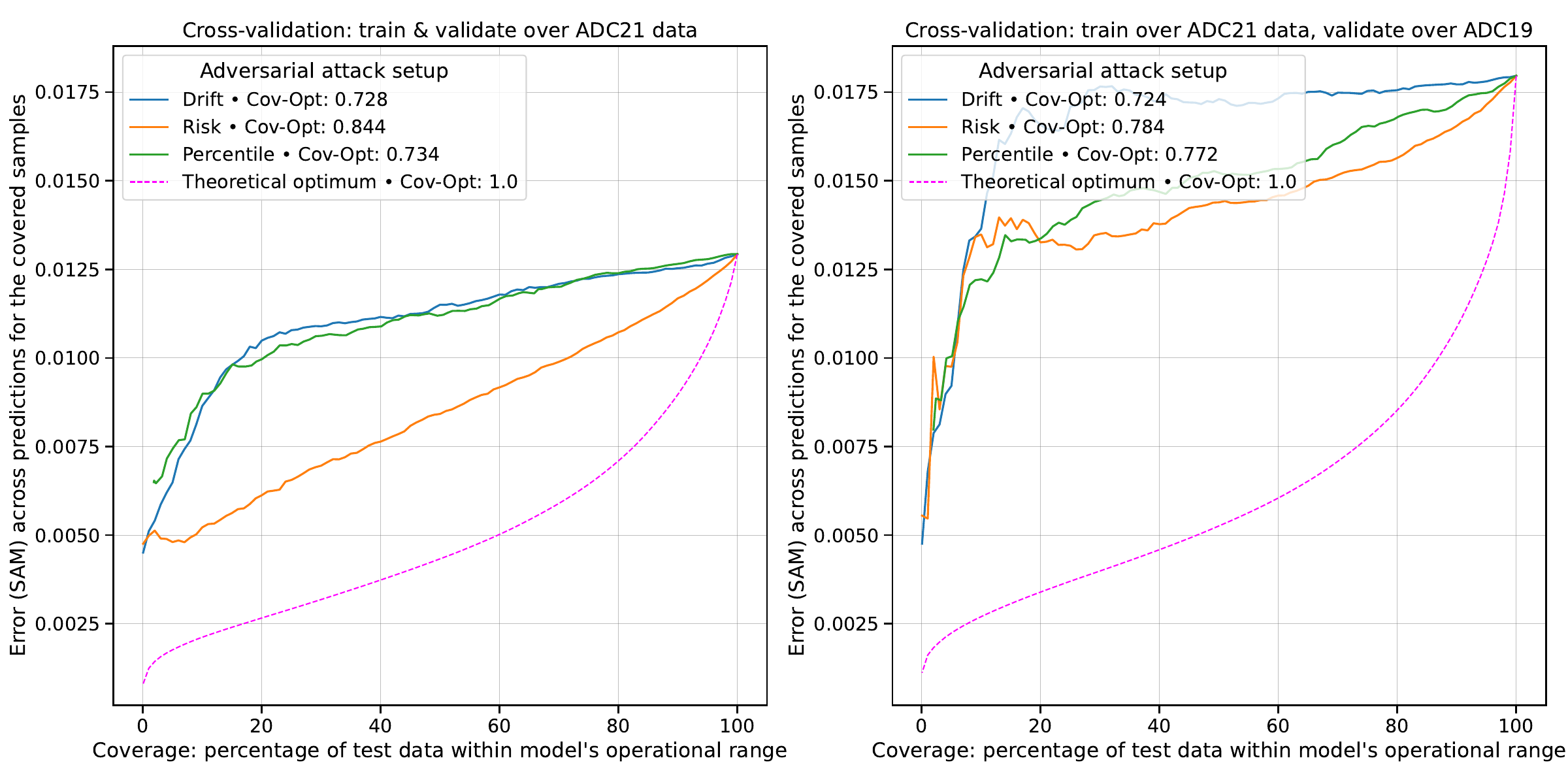}
    \caption{Illustration of SAM evolution as samples are admitted to the model's operational range, based on their safety scores for \textbf{Adversarial Attack} using Drift, Risk, and Percentile, following the approach in \autoref{fig:OODRMSPE}. The x-axis represents the progression from the safest data points to 100\% test data inclusion.}
    \label{fig:AASAM}
\end{figure*}

In the Adversarial Attack setup, three indicators, Drift, Risk, and Percentile, are evaluated to rank the reliability of predictions. Again, the assessment follows the same analytical framework as before.

To simulate operational conditions where astrophysical parameters are not known exactly, a cloud of points is generated around each test input. These perturbations are based on the information provided in the MCS, which contains the error bounds for the different astrophysical parameters. From these bounds, a mean and a standard deviation are obtained for each parameter as a percentage of the corresponding value. A normal distribution is then created for each parameter, and a random combination of parameters is perturbed to see how the model reacts to physically possible variations of the input vector.

The \textbf{Drift} indicator measures the average change in the model's prediction when the input is subjected to this perturbation cloud. The difference between the mean prediction of the cloud and the original prediction is calculated and then divided by the original prediction, obtaining a relative value. The safety score is defined as the negative mean of this absolute relative difference across all wavelengths. A high drift suggests that the model's output is highly dependent on specific parameters, signifying a lack of robustness. The \textbf{Risk} indicator combines both the spread and the shift of the model's response to the perturbation cloud. It incorporates the standard deviation of the cloud predictions and the magnitude of the drift. The final safety score is the negative mean of this combined risk divided by the mean prediction of the cloud across the 55 wavelengths. The \textbf{Percentile} indicator evaluates the symmetry of the perturbed predictions around the original output. It calculates the percentage of cloud predictions that fall below the original point estimate. If the original prediction is robust, it should ideally sit near the median of the perturbed cloud. The safety score penalises cases where the original prediction is an outlier relative to its perturbed neighbours, using the negative mean of 2 x |percentile - 0.5|. A score closer to zero indicates the prediction is central to the distribution of noise-augmented outputs, while a score further from zero indicates fragility.

\begin{table}
    \centering
    \caption{Spearman correlation coefficients between the proposed safety indicators for Adversarial Attack and the prediction error metrics, following the experimental setup in \autoref{tab:OODCorr}.}
    \begin{tabular}{c|c|c|c|c}
    \hline
      Indicator   & RMSPE 21 & RMSPE 19 & SAM 21 & SAM 19\\ \hline
     Drift & -0.234 & -0.209 & -0.178 & -0.167\\
     Risk & -0.414 & -0.391  & -0.505 & -0.479\\
     Percentile &  -0.131 & -0.136 & -0.164 & -0.165 \\ \hline
    \end{tabular}
    \label{tab:AACorr}
\end{table}

The Spearman correlation coefficients in \autoref{tab:AACorr} reveal that Risk is the most effective indicator in this setup. However, in general, these indicators present a weak correlation with the error. The table demonstrates the strongest negative correlations across all categories, reaching $-0.414$ for RMSPE on ADC21 and $-0.505$ for SAM on ADC21. Both Drift and Percentile show significantly weaker relationships with prediction error.

The coverage-error curves in \autoref{fig:AARMSPE} and \autoref{fig:AASAM} justify the selection of Risk as the primary safety indicator. For both RMSPE and SAM tested on ADC21, the Risk indicator maintains a lower and more stable error profile throughout the majority of the coverage range compared to the other indicators and a higher Cov-Opt. When assessed in ADC19, Drift shows better behaviour when evaluated on RMSPE, and Percentile exhibits better initial performance for SAM, although after 20\% coverage, Risk performs better again. In most other instances, Percentile and Drift result in worse coverage curves characterised by high initial errors and lower Cov-Opt scores.

\begin{figure*}
    \centering
    \includegraphics[width=0.8\linewidth]{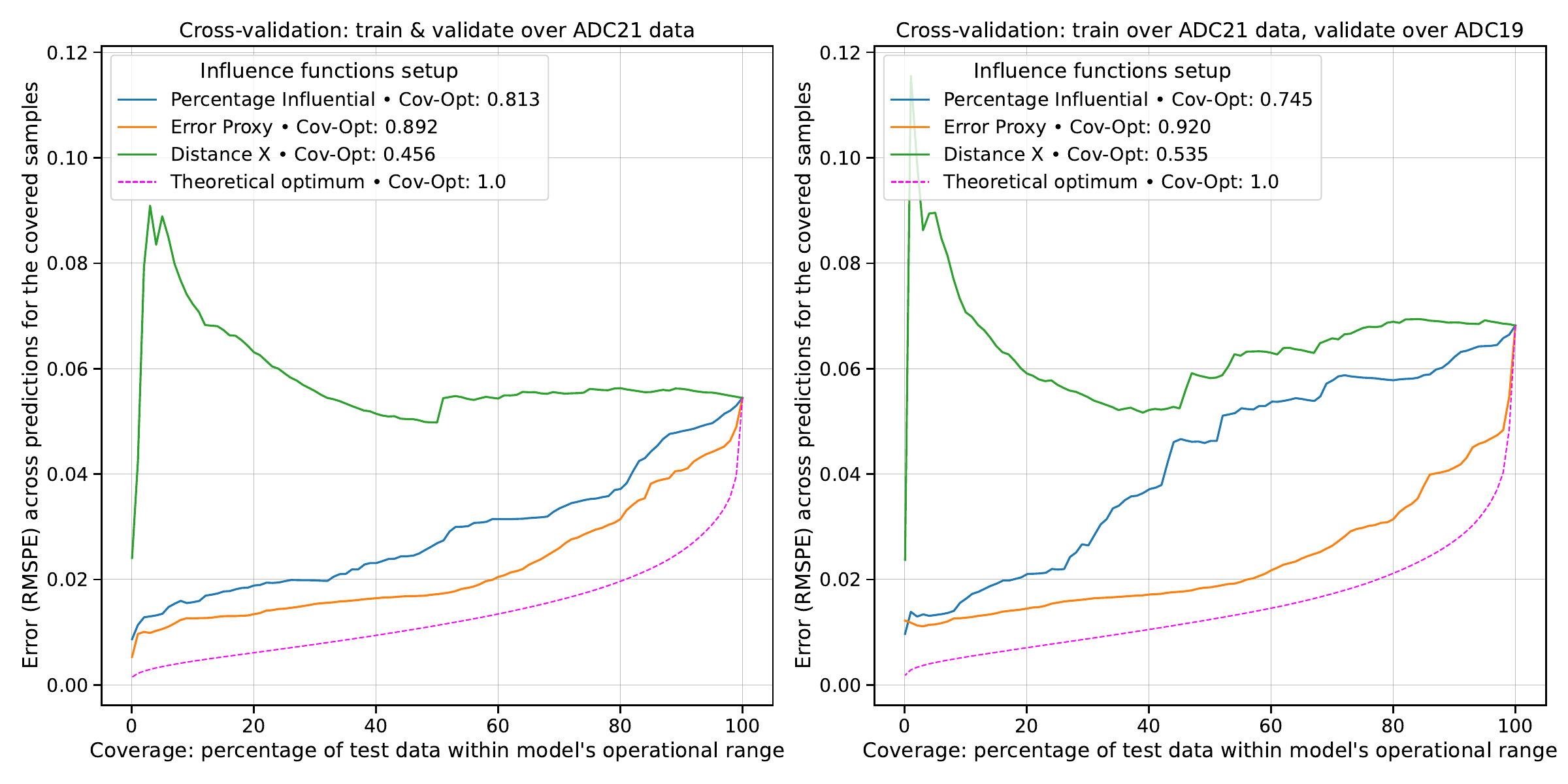}
    \caption{Illustration of RMSPE evolution as samples are admitted to the model's operational range, based on their safety scores for \textbf{Influence Functions} using Percentage Influential, Error Proxy, and Distance $X$, following the approach in \autoref{fig:OODRMSPE}. The x-axis represents the progression from the safest data points to 100\% test data inclusion.}
    \label{fig:IFRMSPE}
\end{figure*}

\begin{figure*}
    \centering
    \includegraphics[width=0.8\linewidth]{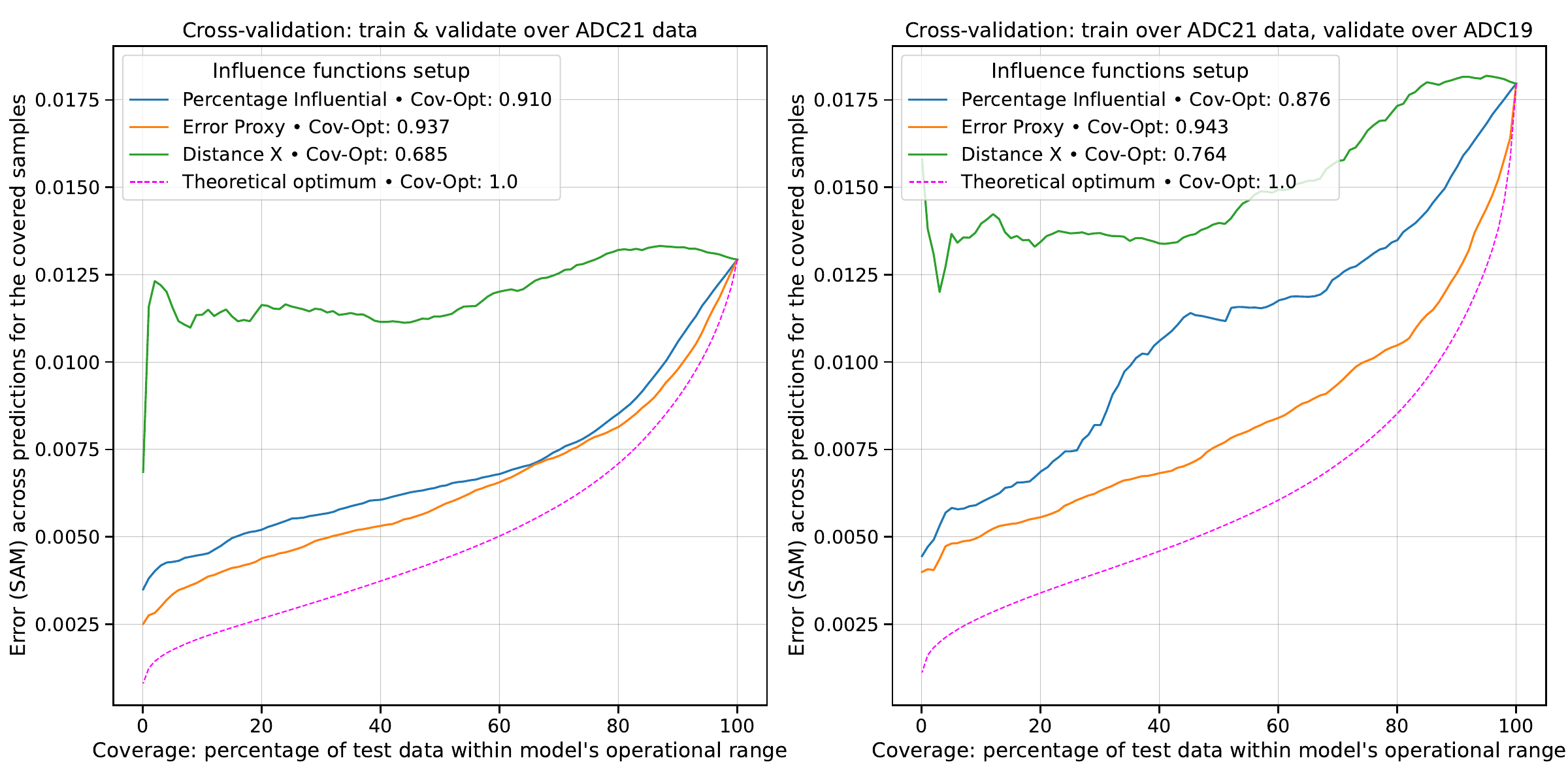}
    \caption{Illustration of SAM evolution as samples are admitted to the model's operational range, based on their safety scores for \textbf{Influence Functions} using Percentage Influential, Error Proxy, and Distance $X$, following the approach in \autoref{fig:OODRMSPE}. The x-axis represents the progression from the safest data points to 100\% test data inclusion.}
    \label{fig:IFSAM}
\end{figure*}

This setup generally shows a degradation in performance when moving from internal validation on ADC21 to cross-domain validation on ADC19, suggesting limited utility in cases involving significant data drift. Because the Risk indicator generally avoids the highest initial errors and has the best Cov-Opt in three out of four cases, it is identified as the most robust metric for this specific setup. Consequently, Risk serves as the primary metric for defining operational safety boundaries in the context of adversarial attacks.

\subsection{Influence Functions}
In the Influence Functions setup, three indicators, Percentage Influential, Error Proxy, and Distance X, are evaluated to rank the reliability of predictions. The assessment follows the same analytical framework as the previous sections, focusing on the correlation between indicators and error metrics across expanding coverage levels. Influence functions assign a signed value to each training point that quantifies how much it pushes a specific prediction up or down, with larger magnitudes indicating a stronger impact and the sign indicating whether the point increases or decreases the prediction in the absence of ground truth \citep{nikkiIF}.

The \textbf{Percentage Influential} indicator measures the concentration of influence. It calculates how many training points are required to reach a specific percentage (e.g., 80\%) of the total influence for a given prediction. The safety score is the ratio of these relevant points to the total training set size. A high score indicates that the prediction is supported by a large group of training examples, while a low score suggests that the prediction relies on only a few influential points, making it more susceptible to errors or outliers. The \textbf{Error Proxy} indicator provides an estimate of the potential prediction error based on the influence of the training set and the Infinitesimal Jackknife estimator variance \citep{jaeckel1972infinitesimal, Jacknifeplusinfluence, swissIJ, nikkiIF}. It calculates the sum of the squared influences for each wavelength, normalises them by the square of the model's prediction, and takes the root mean across all wavelengths. The resulting safety score is the negative of this estimated error. This metric serves as a proxy for model confidence because if a prediction is driven by a set of mismodelled training points, the estimated error will be higher, resulting in a lower safety score. The \textbf{Distance X} indicator evaluates the geometric similarity between a test sample and its most influential training point. By identifying the specific training example that had the greatest impact on the current prediction, the safety model calculates the RMSPE distance between the two input vectors. The safety score is the negative of this distance. This operates on the idea that if the model's prediction is primarily influenced by a training point that is physically very different from the current input, the prediction is considered less safe.

\begin{table}
    \centering
    \caption{Spearman correlation coefficients between the proposed safety indicators for Influence Functions and the prediction error metrics, following the experimental setup in \autoref{tab:OODCorr}.}
    \begin{tabular}{c|c|c|c|c}
    \hline
      Indicator   & RMSPE 21 & RMSPE 19 & SAM 21 & SAM 19\\ \hline
     Percentage Influential  & -0.578 & -0.575 & -0.612 & -0.540\\
     Error Proxy  & -0.662  & -0.670 & -0.724 & -0.678\\
     Distance X &  0.041 &  0.018 & -0.032 & -0.081 \\ \hline
    \end{tabular}
    \label{tab:IFCorr}
\end{table}

The Spearman correlation coefficients in \autoref{tab:IFCorr} demonstrate that Error Proxy is the most effective indicator for this setup. It exhibits the strongest negative correlations across all categories, reaching $-0.662$ for RMSPE on ADC21 and $-0.724$ for SAM on ADC21. Percentage Influential also shows strong performance, with correlations ranging between $-0.540$ and $-0.612$. Conversely, Distance X proves to be an unreliable indicator for this problem, as it shows nearly zero or slightly positive correlations, reaching only $-0.081$ for SAM on ADC19. This suggests that the distance in the input space does not reliably correspond to prediction error in this context.

The coverage-error curves in \autoref{fig:IFRMSPE} and \autoref{fig:IFSAM} justify the selection of Error Proxy as the primary safety indicator. For both RMSPE and SAM, the Error Proxy indicator achieves the highest Cov-Opt score across all cases and maintains the lowest and most stable error profile throughout the coverage range compared to the other indicators. While Percentage Influential displays a similar trend, it generally maintains a higher error plateau as coverage increases. The Distance X indicator consistently results in the highest error levels from the start, showing almost no ability to distinguish between high and low error samples at low coverage levels. The performance of the Error Proxy indicator remains remarkably stable when moving from internal validation on ADC21 to cross-domain validation on ADC19.

Because the Error Proxy indicator avoids the high initial errors and poor ranking associated with Distance X, it is identified as the most robust metric for this setup. This setup appears particularly promising for identifying safe operational regions, as it maintains a high degree of predictive power even under domain shift, and achieves higher Cov-Opts in the cross-domain scenarios.

\subsection{Activation Pattern Monitoring}
\begin{figure*}
    \centering
    \includegraphics[width=0.8\linewidth]{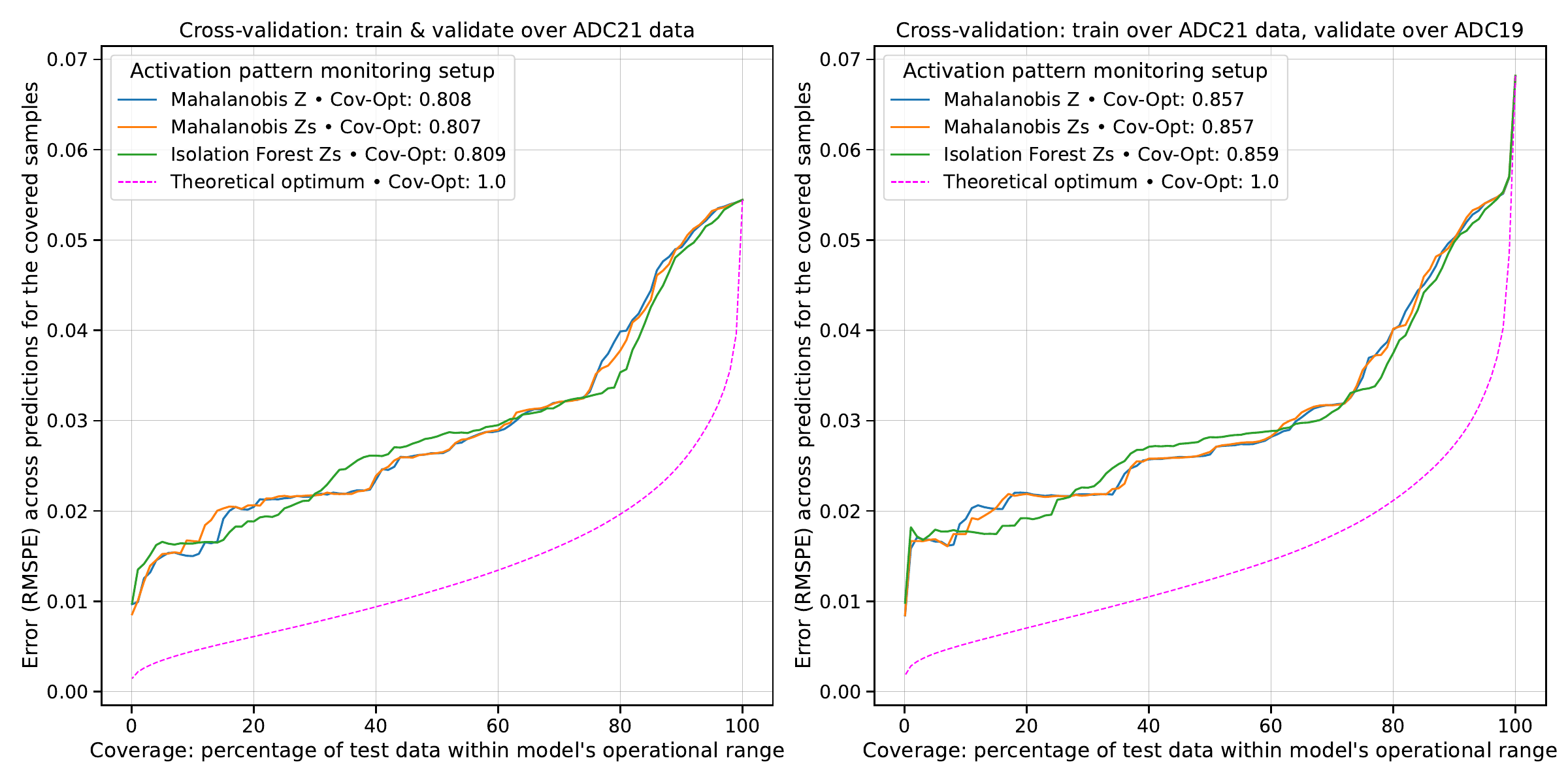}
    \caption{Illustration of RMSPE evolution as samples are admitted to the model's operational range, based on their safety scores for \textbf{Activation Pattern Monitoring} using Mahalanobis Z, Mahalanobis Zs, and Isolation Forest Zs, following the approach in \autoref{fig:OODRMSPE}. The x-axis represents the progression from the safest data points to 100\% test data inclusion.}
    \label{fig:APRMSPE}
\end{figure*}

\begin{figure*}
    \centering
    \includegraphics[width=0.8\linewidth]{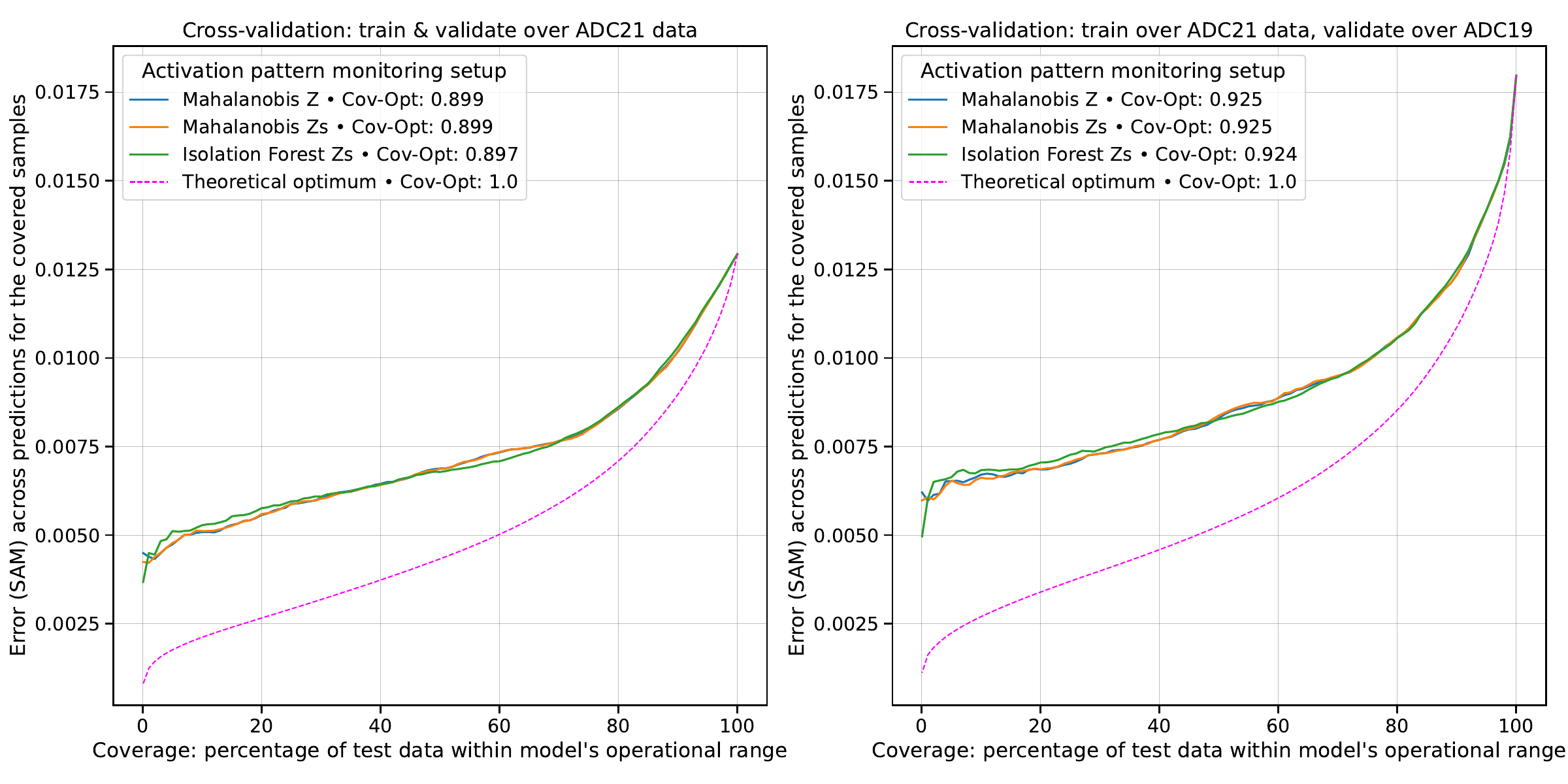}
    \caption{Illustration of SAM evolution as samples are admitted to the model's operational range, based on their safety scores for \textbf{Activation Pattern Monitoring} using Mahalanobis Z, Mahalanobis Zs, and Isolation Forest Zs, following the approach in \autoref{fig:OODRMSPE}. The x-axis represents the progression from the safest data points to 100\% test data inclusion.}
    \label{fig:APSAM}
\end{figure*}

In the Activation Pattern Monitoring setup, three indicators, Mahalanobis Z, Mahalanobis Zs, and Isolation Forest Zs, are evaluated to rank the reliability of predictions.

The framework extracts internal representations from the model, specifically focusing on the Z-layer, the linear combination preceding the activation layer. These representations are employed in two forms: raw and standardised. While Z corresponds to the raw hidden layer activations, Zs represents the standardised activations, where each dimension is scaled according to the mean and standard deviation observed for that specific hidden layer during the training phase.

The \textbf{Mahalanobis Z} and \textbf{Mahalanobis Zs} indicators calculate the Mahalanobis distance between the current internal activation pattern and the average pattern established during training \citep{mahalanobis1936distance}. To manage the high dimensionality of the hidden layer and balance computational requirements, PCA is first applied to retain only the first 50 components. Unlike Euclidean distance, the Mahalanobis distance accounts for correlations between different hidden units. A larger distance indicates that the model is in an unusual internal state, suggesting the prediction may be unreliable. The safety score is thus defined as the negative of this distance, ensuring that higher scores represent safer predictions. The \textbf{Isolation Forest Zs} indicator applies an anomaly detection model directly to the standardised activation patterns. This follows the same principle as OOD detection, but is specifically applied to the standardised Z-layer. If a test sample triggers a sequence of internal activations that occurred rarely during training, the Isolation Forest assigns it a lower decision score, thereby flagging it as an out-of-domain internal state.

\begin{table}
    \centering
    \caption{Spearman correlation coefficients between the proposed safety indicators for Activation Pattern Monitoring and the prediction error metrics, following the experimental setup in \autoref{tab:OODCorr}.}
    \begin{tabular}{c|c|c|c|c}
    \hline
      Indicator   & RMSPE 21 & RMSPE 19 & SAM 21 & SAM 19\\ \hline
     Mahalanobis Z & -0.556 & -0.583 & -0.580 & -0.572\\
     Mahalanobis Zs   &  -0.556 & -0.584 & -0.581 & -0.572\\
     Isolation Forest Zs &  -0.539 & -0.563 &  -0.579 & -0.569\\ \hline
    \end{tabular}
    \label{tab:APCorr}
\end{table}

The Spearman correlation coefficients in \autoref{tab:APCorr} show that all three indicators demonstrate remarkably similar and strong performance. Mahalanobis Z and Mahalanobis Zs exhibit nearly identical results, reaching correlations of approximately $-0.57$ for both RMSPE and SAM. Isolation Forest Zs follows closely, with correlation values only slightly lower. These highly consistent negative correlations across both internal validation (ADC21) and cross-domain validation (ADC19) suggest that activation-based monitoring is highly robust to distributional shifts.

The coverage-error curves in \autoref{fig:APRMSPE} and \autoref{fig:APSAM} further illustrate the consistent performance of these indicators. For both RMSPE and SAM, the three curves overlap significantly, maintaining a low and stable error profile as coverage increases from the safest data points. Unlike other setups where specific indicators misidentified initial safe cases, the activation-based metrics correctly prioritise low-error samples from the onset. This stability is maintained in the cross-domain validation on ADC19, where the indicators effectively manage the admission of data despite the presence of drift.

Because all three indicators provide nearly identical and reliable trade-offs between maximising data coverage and minimising prediction error, they are all identified as robust choices for this setup. However, Mahalanobis Zs is selected as the primary metric for defining operational safety boundaries. This choice is based on its slight statistical advantage in correlation and its ability to provide a standardised score that remains stable across different datasets, making it highly suitable for real-world deployment where consistent safety thresholds are required.

\section{Plots for Combinations and Aggregation Analysis}

The remainder of the plots discussed in \autoref{sub:ICAA} are shown below.

\begin{figure*}
    \centering
    \begin{subfigure}{0.48\textwidth}
        \hspace{-12mm}
        \includegraphics[width=1.2\linewidth]{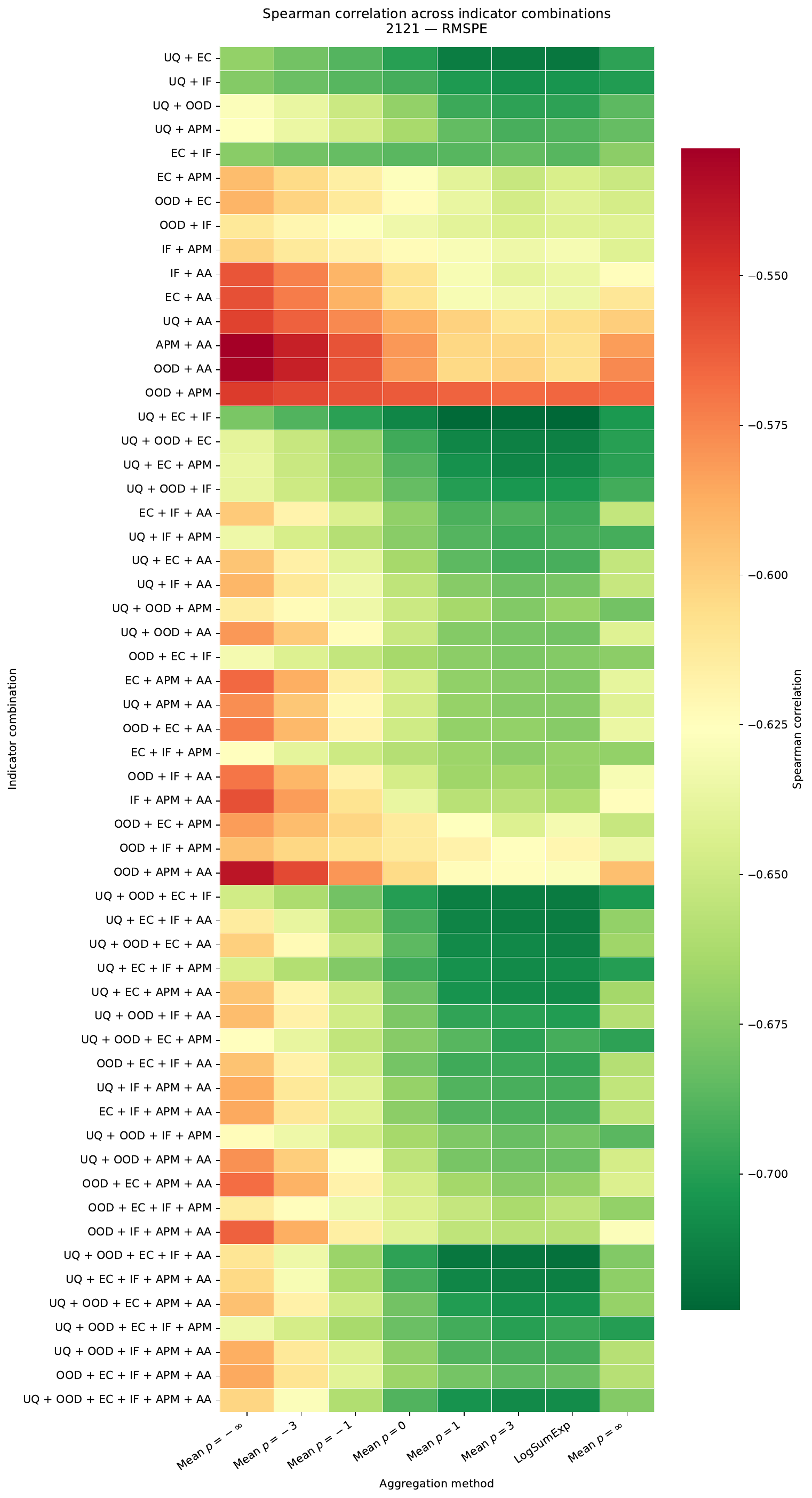}
        \caption{ADC21 $\rightarrow$ ADC21}
        \label{fig:heatmap21RMSPE}
    \end{subfigure}
    \hfill
    \begin{subfigure}{0.48\textwidth}
        \hspace{-5mm}
        \includegraphics[width=1.2\linewidth]{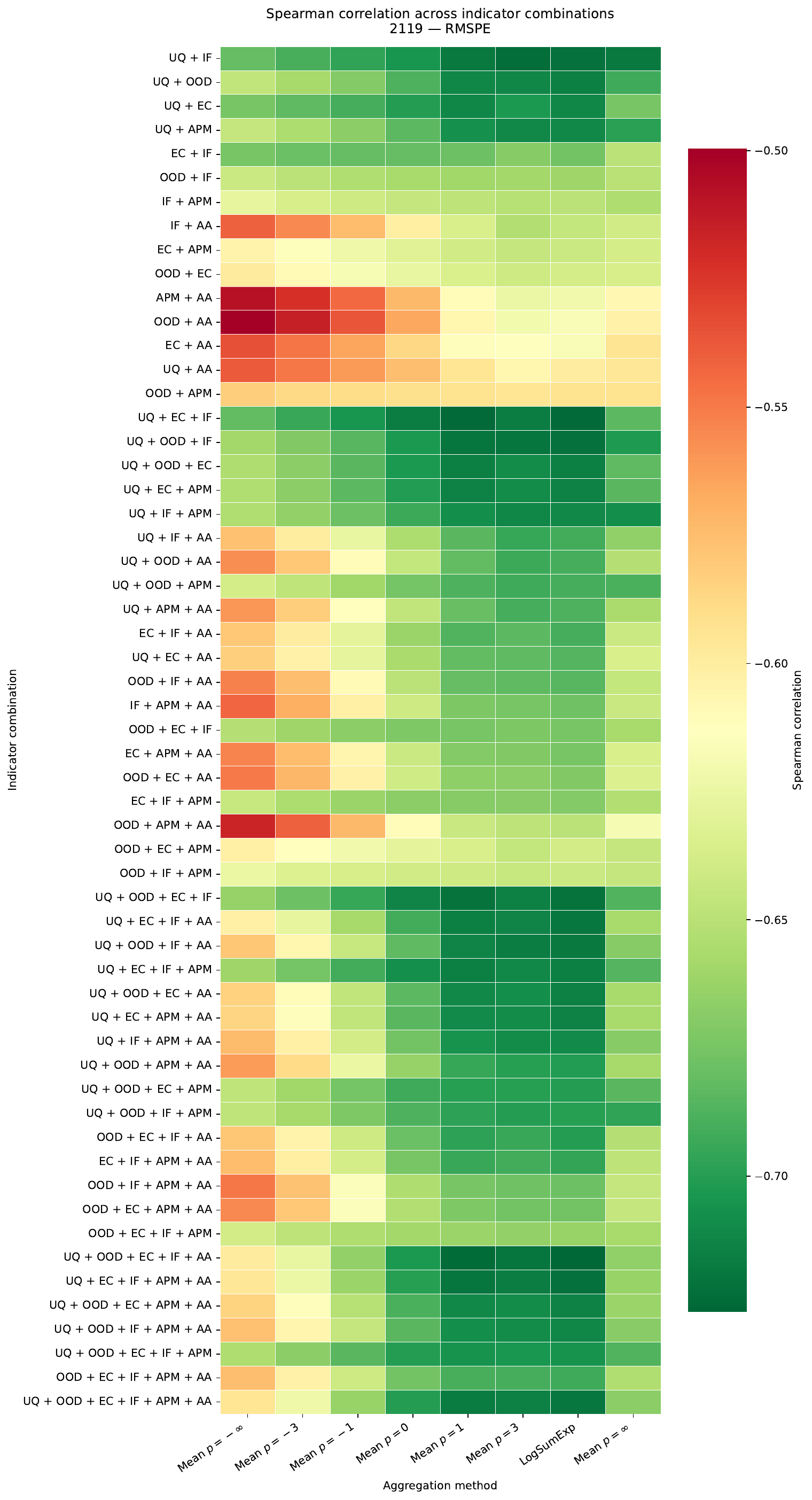}
        \caption{ADC21 $\rightarrow$ ADC19}
        \label{fig:heatmap19RMSPE}
    \end{subfigure}

    \caption{Spearman correlation between aggregated risk scores and RMSPE across all indicator combinations and aggregation methods. Left: training and testing on ADC21. Right: training on ADC21 and testing on ADC19. More negative values indicate better alignment between the risk indicator and true error.}
\end{figure*}

\begin{figure*}
    \centering
    \begin{subfigure}{0.48\textwidth}
        \hspace{-12mm}
        \includegraphics[width=1.2\linewidth]{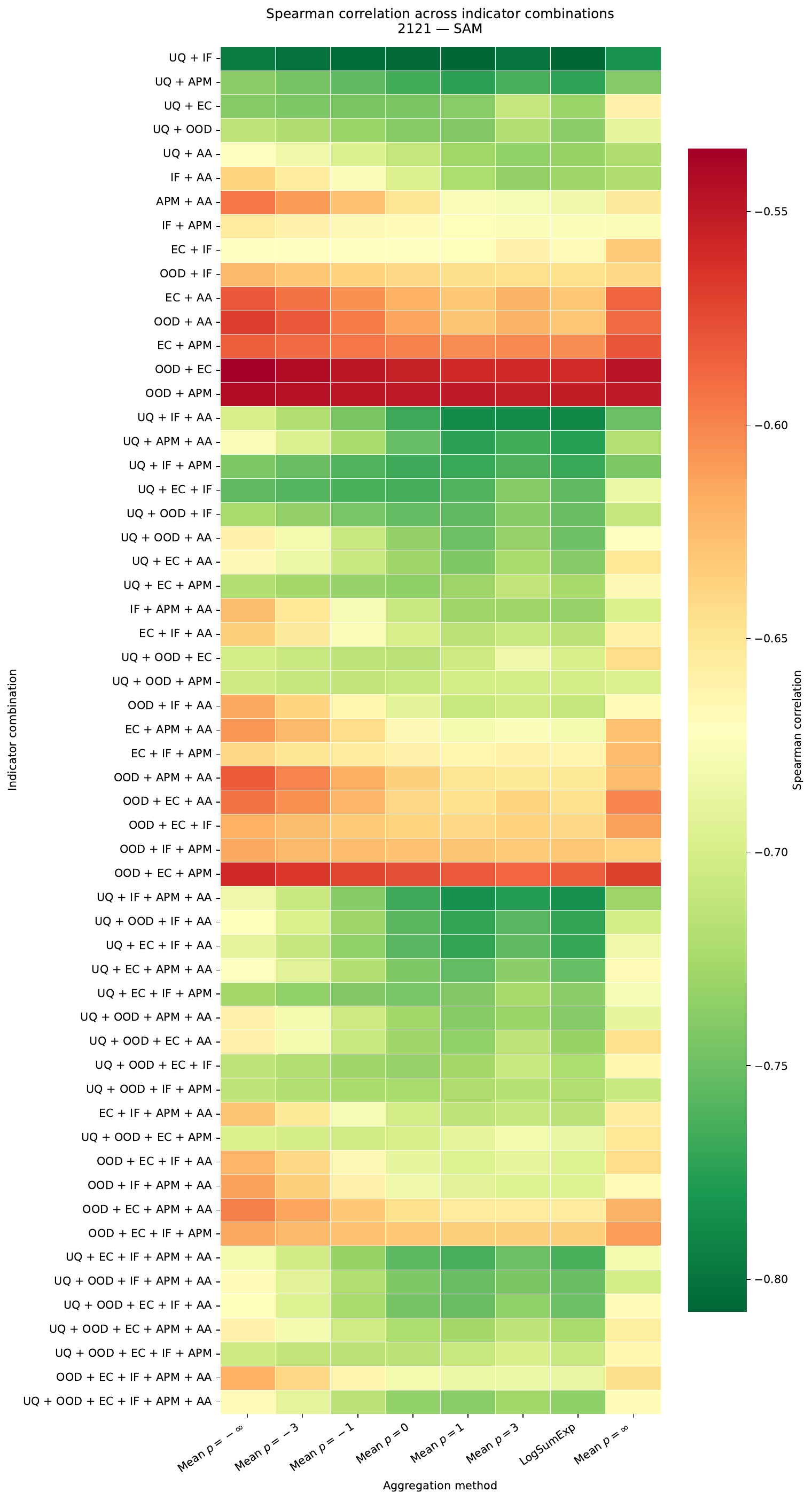}
        \caption{ADC21 $\rightarrow$ ADC21}
        \label{fig:heatmap21SAM}
    \end{subfigure}
    \hfill
    \begin{subfigure}{0.48\textwidth}
        \hspace{-5mm}
        \includegraphics[width=1.2\linewidth]{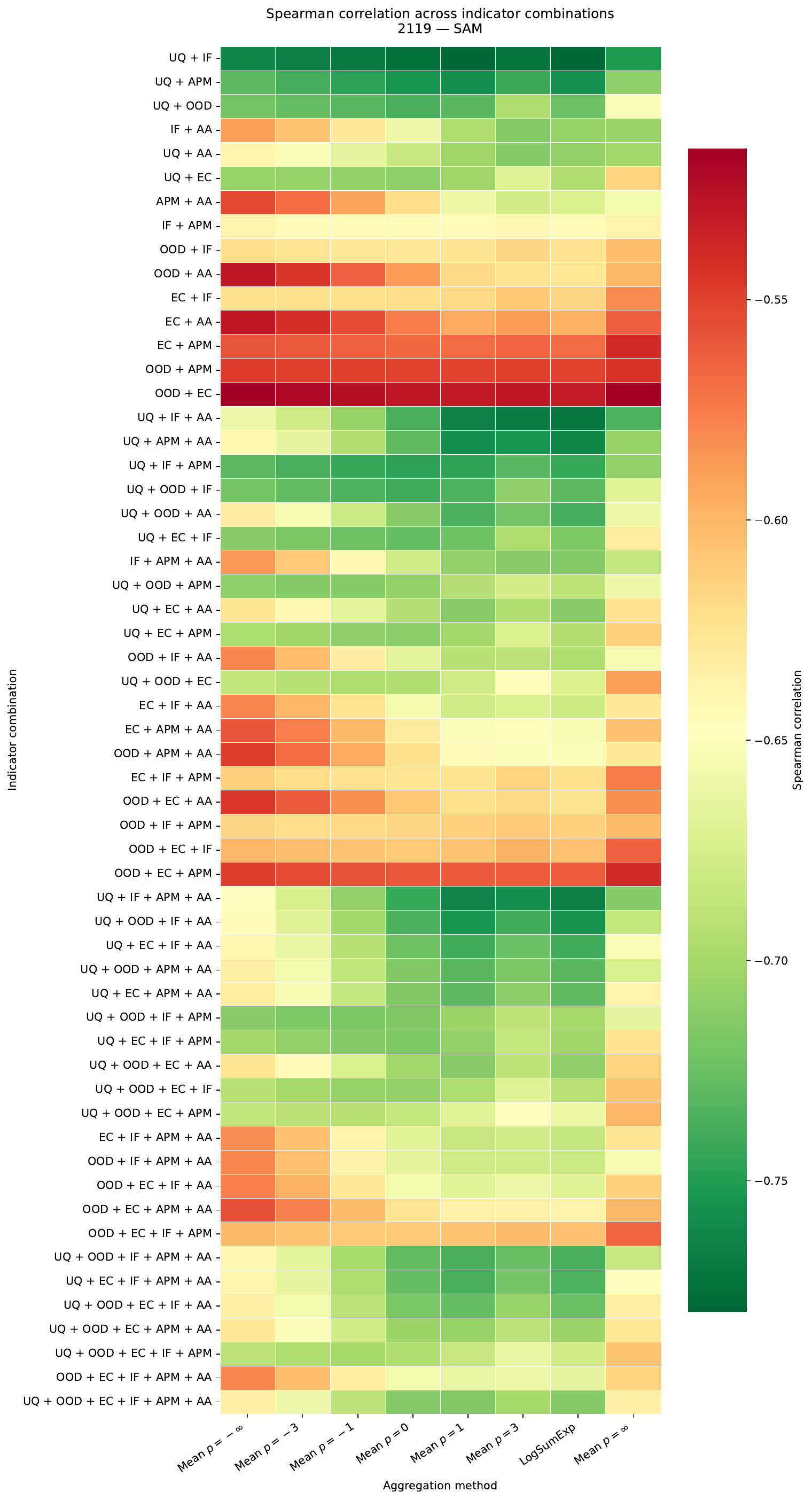}
        \caption{ADC21 $\rightarrow$ ADC19}
        \label{fig:heatmap19SAM}
    \end{subfigure}

    \caption{Spearman correlation between aggregated risk scores and SAM across all indicator combinations and aggregation methods. Left: training and testing on ADC21. Right: training on ADC21 and testing on ADC19. More negative values indicate better alignment between the risk indicator and true error.}
\end{figure*}

\begin{figure*}
    \centering
    \includegraphics[width=0.7\linewidth]{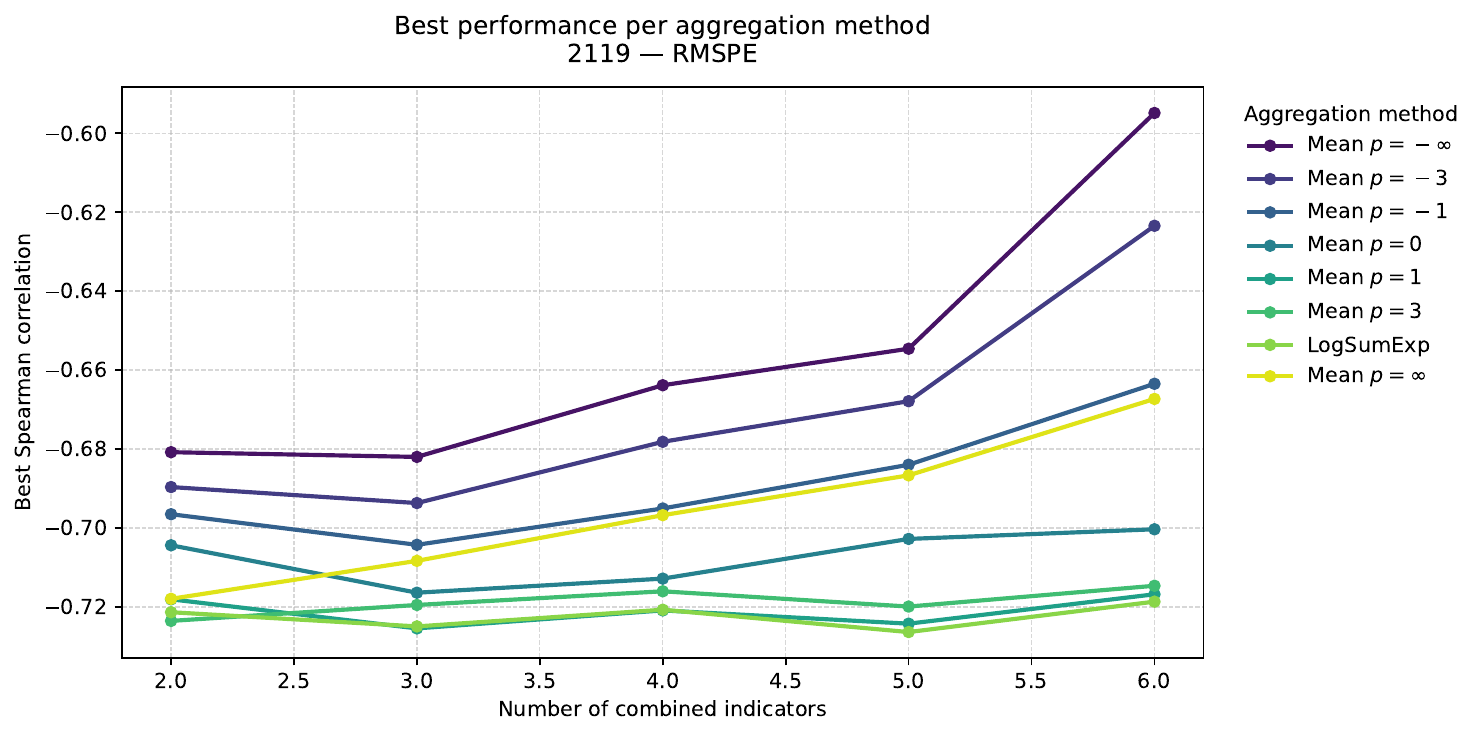}
    \caption{Best Spearman correlation per aggregation method as a function of indicator combination size for RMSPE when training on ADC21 and testing on ADC19.}
    \label{fig:BestSpearman19RMSPE}
\end{figure*}

\begin{figure*}
    \centering
    \includegraphics[width=0.7\linewidth]{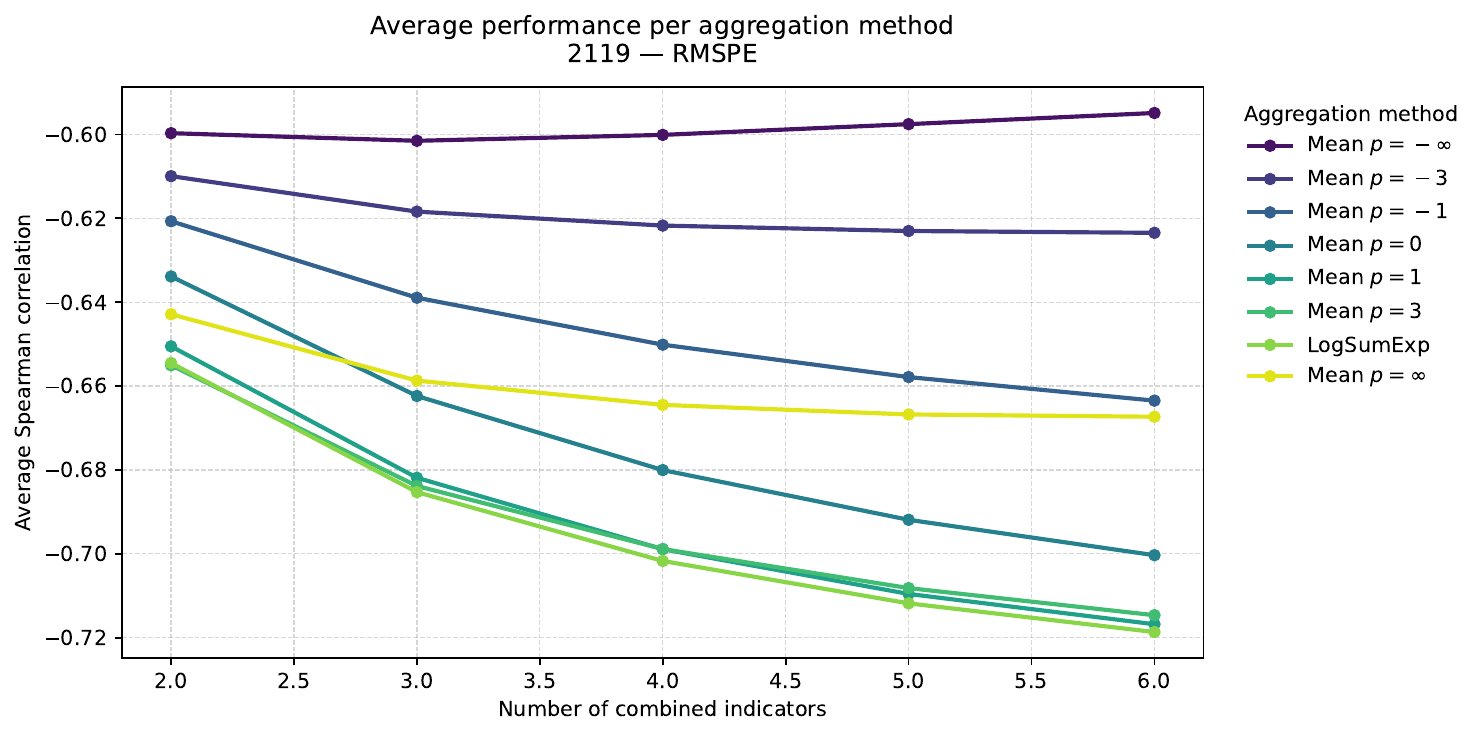}
    \caption{Average Spearman correlation per aggregation method as a function of indicator combination size for RMSPE training on ADC21 and testing on ADC19.}
    \label{fig:AverageSpearman19RMSPE}
\end{figure*}

\begin{figure*}
    \centering
    \includegraphics[width=0.7\linewidth]{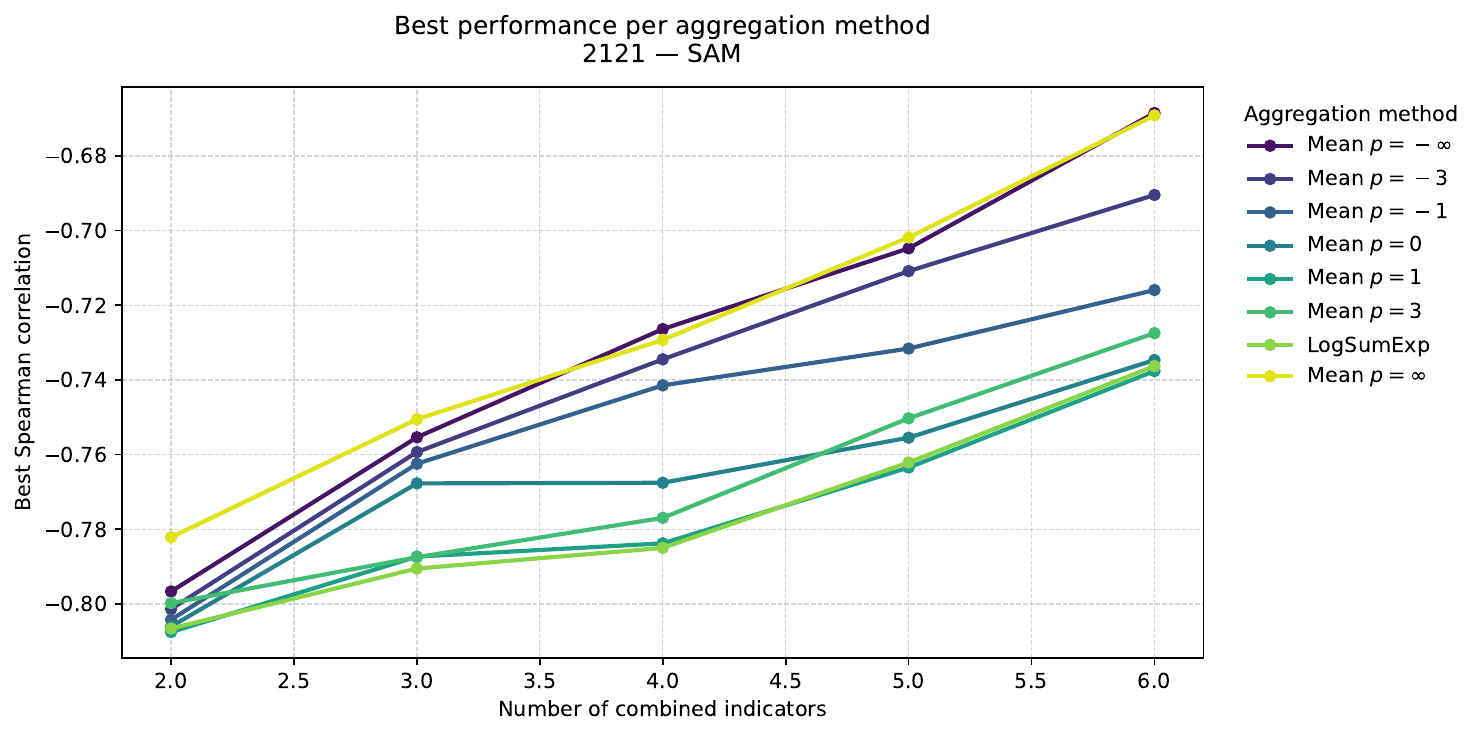}
    \caption{Best Spearman correlation per aggregation method as a function of indicator combination size for SAM when training and testing on ADC21. More negative values indicate better alignment between the risk indicator and true error.}
    \label{fig:BestSpearman21SAM}
\end{figure*}

\begin{figure*}
    \centering
    \includegraphics[width=0.7\linewidth]{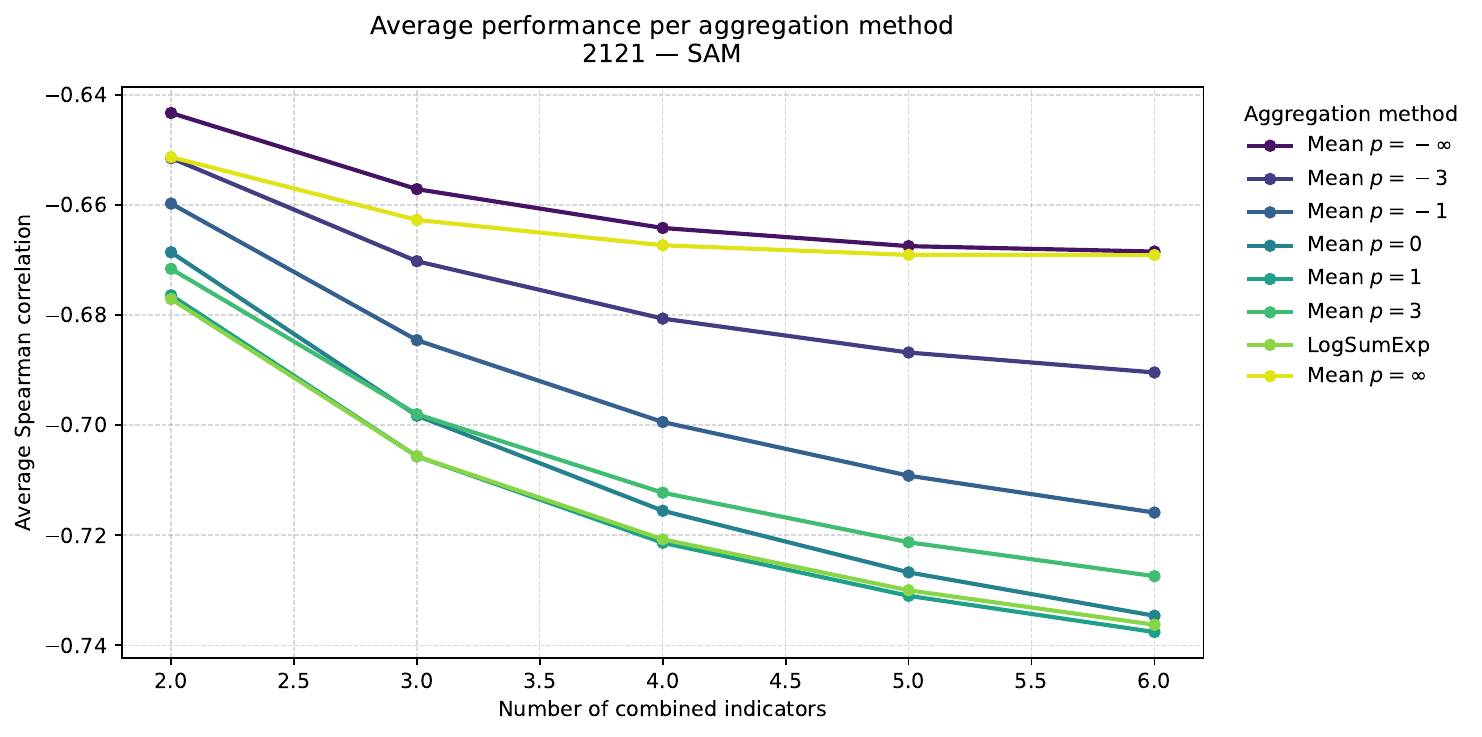}
    \caption{Average Spearman correlation per aggregation method as a function of indicator combination size for SAM when training and testing on ADC21.}
    \label{fig:AverageSpearman21SAM}
\end{figure*}

\begin{figure*}
    \centering
    \includegraphics[width=0.7\linewidth]{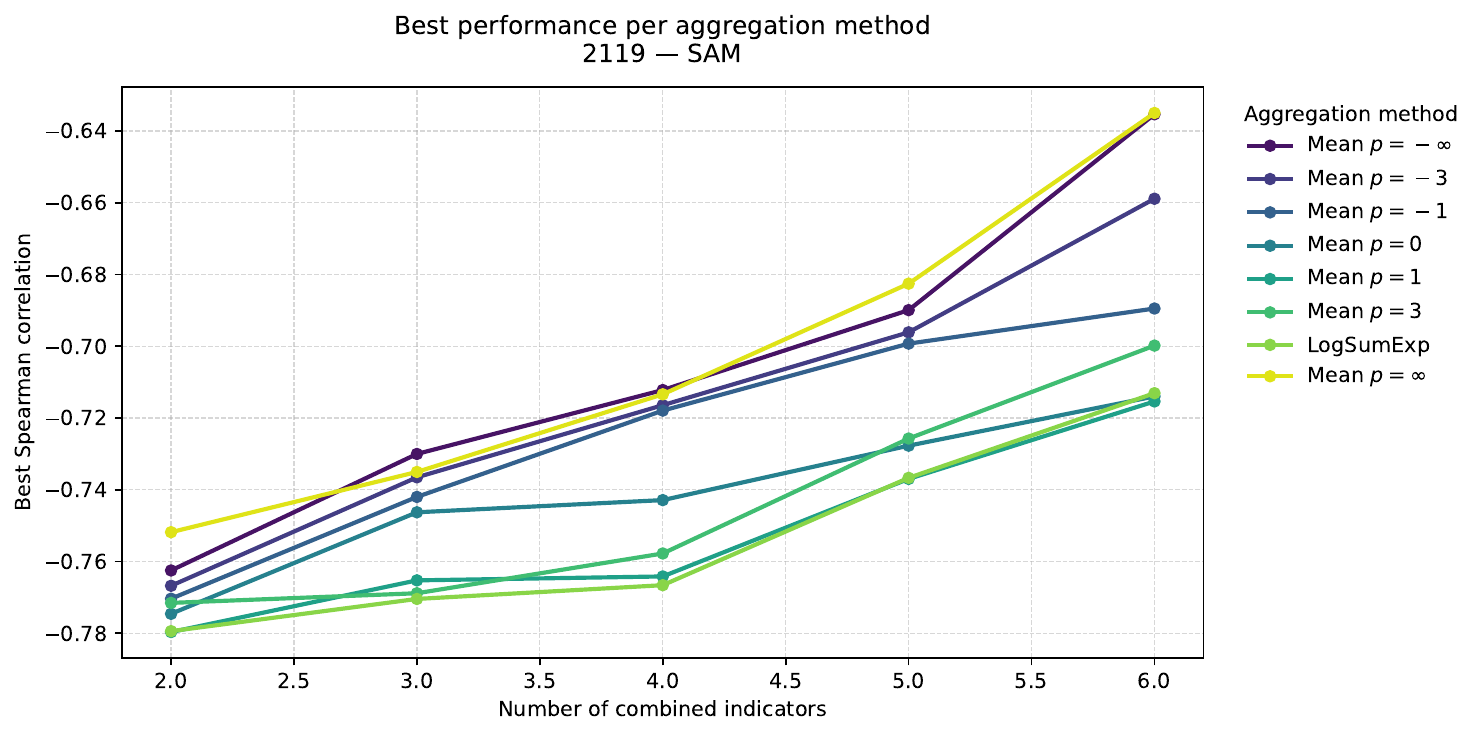}
    \caption{Best Spearman correlation per aggregation method as a function of indicator combination size for SAM when training on ADC21 and testing on ADC19.}
    \label{fig:BestSpearman19SAM}
\end{figure*}

\begin{figure*}
    \centering
    \includegraphics[width=0.7\linewidth]{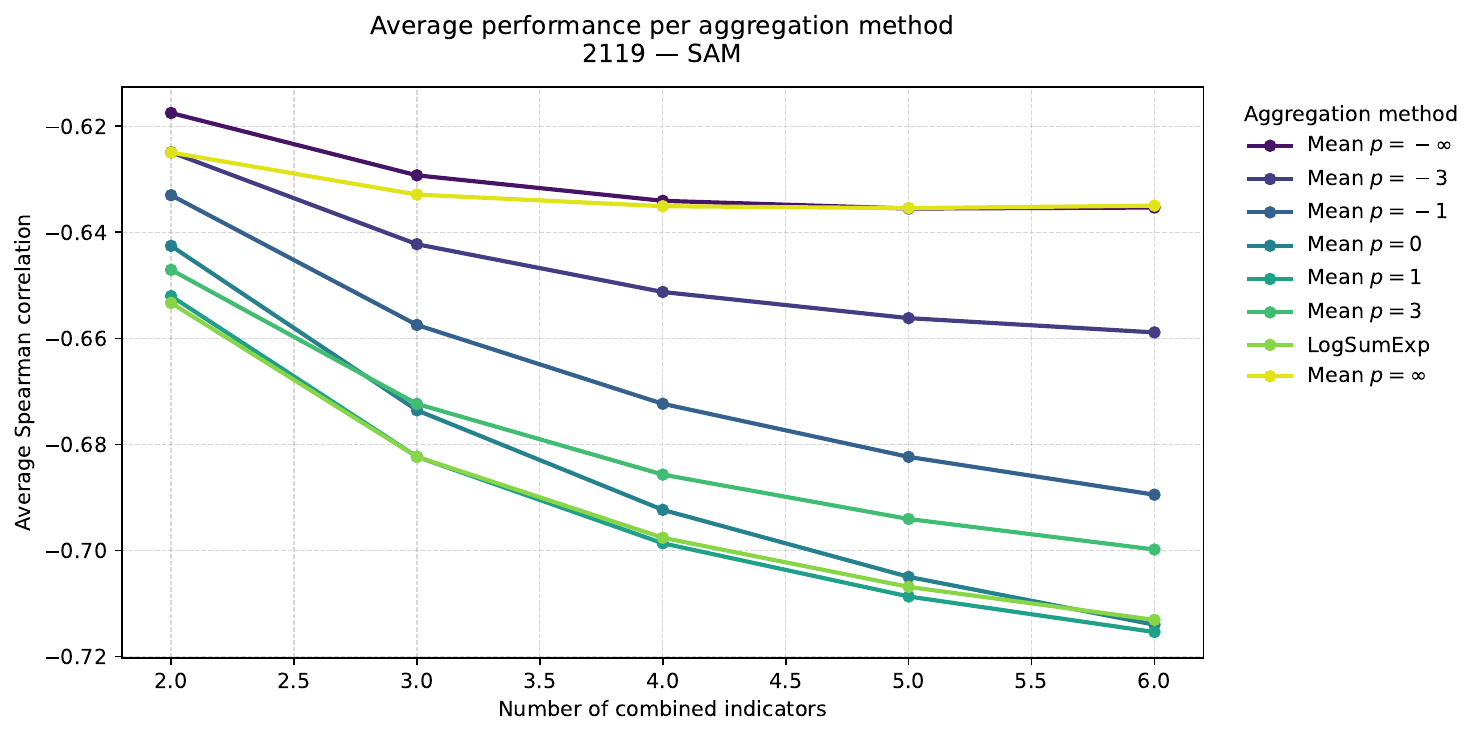}
    \caption{Average Spearman correlation per aggregation method as a function of indicator combination size for SAM when training on ADC21 and testing on ADC19.}
    \label{fig:AverageSpearman19SAM}
\end{figure*}


\bsp	
\label{lastpage}
\end{document}